%% file: main.tex
\documentclass[10pt,twocolumn,letterpaper]{article}

\usepackage{cvpr}              

\input{preamble}

\definecolor{cvprblue}{rgb}{0.21,0.49,0.74}
\usepackage[pagebackref,breaklinks,colorlinks,citecolor=cvprblue]{hyperref}

\usepackage{booktabs}
\usepackage{color}
\usepackage{multirow}
\usepackage{amsthm}
\usepackage{tabularx}
\usepackage{enumerate}
\usepackage[accsupp]{axessibility} 
\usepackage{enumitem}

\def\paperID{11223} 
\def\confName{CVPR}
\def\confYear{2024}

\title{Exploring Efficient Asymmetric Blind-Spots for Self-Supervised Denoising in
Real-World Scenarios}

\author{
Shiyan Chen$^{1,2}$\quad Jiyuan Zhang$^{1,2}$\quad Zhaofei Yu$^{1,2,3}$\thanks{Corresponding author.}\quad Tiejun Huang$^{1,2,3}$  \\
$^1$School of Computer Science, Peking University\\
$^2$National Key Laboratory for Multimedia Information Processing, Peking University\\
$^3$Institute for Artificial Intelligence, Peking University\\
{\tt\small \{strerichia002p,jyzhang\}@stu.pku.edu.cn}, {\tt\small\{yuzf12,tjhuang\}@pku.edu.cn} \\
}

\begin{document}
\maketitle

\input{sec/0_abstract}    
\input{sec/1_intro}
\input{sec/2_related}

\input{sec/3_method}

\input{sec/4_experiment}

\input{sec/5_conclusion}
{
    \small
    \bibliographystyle{ieeenat_fullname}
    \bibliography{main}
}

\input{sec/X_suppl}

\end{document}

%% file: preamble.tex
\usepackage[dvipsnames]{xcolor}


%% file: sec/0_abstract.tex
\begin{abstract}

     Self-supervised denoising has attracted widespread attention due to its ability to train without clean images. However, noise in real-world scenarios is often spatially correlated, which causes many self-supervised algorithms that assume pixel-wise independent noise to perform poorly. Recent works have attempted to break noise correlation with downsampling or neighborhood masking. However, denoising on downsampled subgraphs can lead to aliasing effects and loss of details due to a lower sampling rate. Furthermore, the neighborhood masking methods either come with high computational complexity or do not consider local spatial preservation during inference. Through the analysis of existing methods, we point out that the key to obtaining high-quality and texture-rich results in real-world self-supervised denoising tasks is to train at the original input resolution structure and use asymmetric operations during training and inference. Based on this, we propose Asymmetric Tunable Blind-Spot Network (AT-BSN), where the blind-spot size can be freely adjusted, thus better balancing noise correlation suppression and image local spatial destruction during training and inference. In addition, we regard the pre-trained AT-BSN as a meta-teacher network capable of generating various teacher networks by sampling different blind-spots. We propose a blind-spot based multi-teacher distillation strategy to distill a lightweight network, significantly improving performance. Experimental results on multiple datasets prove that our method achieves state-of-the-art, and is superior to other self-supervised algorithms in terms of computational overhead and visual effects.

\end{abstract}

%% file: sec/1_intro.tex
\section{Introduction}
\label{sec:intro}

Image denoising is an essential low-level computer vision problem. With the advancements in deep learning, an increasing number of studies are focused on supervised learning using clean-noisy pairs~\cite{RIDNet,CBDNet,FADNet,DANet,DnCNN,FFDNet}. Typically, additive white Gaussian noise (AWGN) is introduced into clean datasets to synthesize clean-noisy denoising datasets. However, real-world noise is known to be spatially correlated~\cite{chatterjee2011noise,jin2020review,park2009case}.
Some generative-based methods attempt to synthesize real-world noise from existing clean data~\cite{GAN2GAN,GCBD,UIDNet,C2N,DBSN}. However, synthesizing real-world noise remains challenging, and suffers from generalization issues.
To address the issue, some researchers attempt to capture clean-noisy pairs in real-world scenarios~\cite{SIDD,NIND}. However, in certain scenarios, such as medical imaging and electron microscopy, constructing such datasets can be impractical or even infeasible. 

\begin{figure}
    \centering
    \includegraphics[width=1\linewidth]{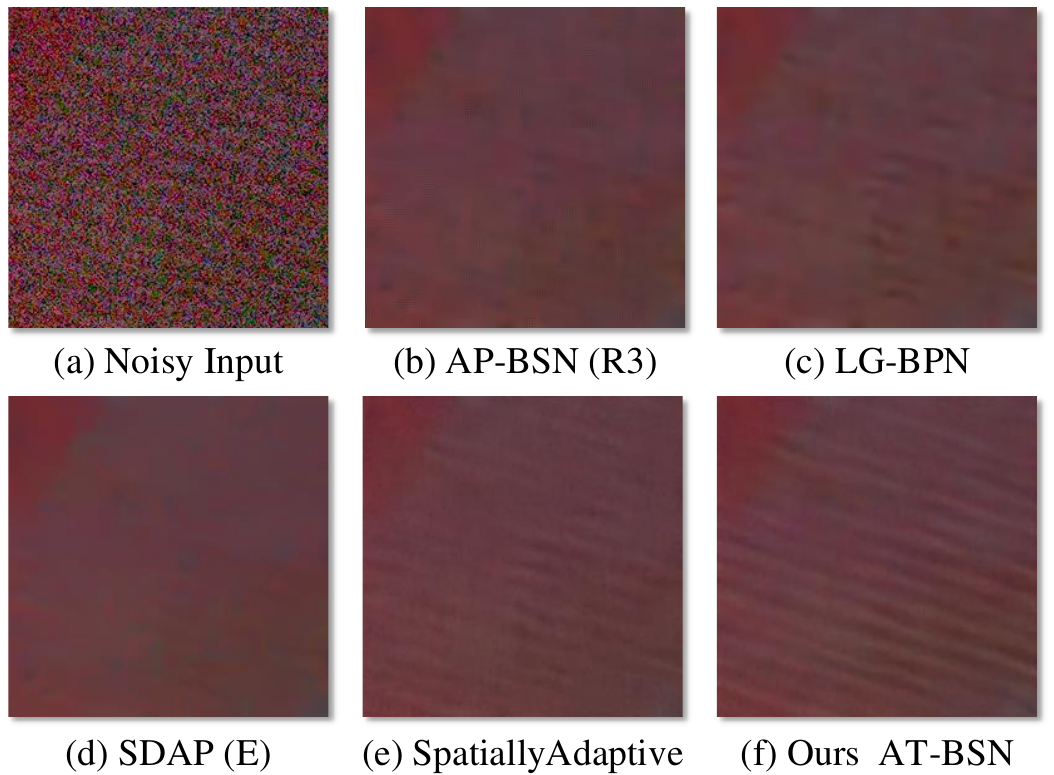}
    \caption{Comparisons of our AT-BSN with other methods. Our method recovers more high frequency texture details.}
    \vspace{-3mm}
    \label{fig:head}
\end{figure}

Self-supervised denoising algorithms, represented by Noise2Noise~\cite{Noise2Noise}, have brought new life to the denoising field.  These methods only require noisy observations to train the denoising model. However, in real-world scenarios, noise often exhibits spatial correlation, which contradicts the pixel-wise independent noise assumption~\cite{Noise2Void,Noise2Noise} that most self-supervised algorithms~\cite{Noise2Self,Neighbor2Neighbor,Noise2Void,Noise2Noise,Blind2Unblind} rely on. Recent studies have proposed self-supervised denoising algorithms suitable for real-world scenarios~\cite{APBSN,CVFSID,li2023spatially,wang2023lg,pan2023random,jang2023self}. These methods mainly disrupt the noise correlation by downsampling~\cite{APBSN,pan2023random,jang2023self} or neighborhood masking~\cite{li2023spatially,wang2023lg}. 
The representative work of the former is AP-BSN~\cite{APBSN}, which utilized pixel-shuffle downsampling (PD)~\cite{APBSN,whenAWGN} to disrupt the noise correlation and employed asymmetric PD stride factors for training and inference. 
However, according to the Nyquist-Shannon sampling theorem, downsampling methods disrupt image spatial structure during inference, leading to lower sampling density and loss of high-frequency details. Conversely, neighborhood masking methods denoise at original resolution structure, retaining more texture information.

In this paper, we carefully analyze the existing related work and point out that \textbf{training at the original resolution structure} and \textbf{using asymmetric operations during training and inference} are key to producing high-quality, texture-rich clear images in self-supervised real noise removal tasks. Based on these observations, we propose a novel paradigm called \textbf{Asymmetric Tunable Blind-Spot Network(AT-BSN)}, where the blind-spot size can be freely adjusted to balance between noise correlation suppression and image local structure destruction. Furthermore, the flexible tunable blind-spot allows us to obtain a potential teacher network distribution, where each sampled teacher has a different blind-spot, making each teacher network’s ability to handle flat/texture areas different. We propose a \textbf{Blind-Spots Based Multi-Teacher Distillation} strategy, which significantly improves performance and further reduces computational overhead. 
Experimental results demonstrate the effectiveness of the proposed method.

The main contributions are summarized as follows:
\begin{itemize}
    \item We carefully analyze existing methods and point out that training at the original resolution structure and using asymmetric operations during training and inference are key to producing high-quality, texture-rich results in self-supervised denoising tasks.
    \item We propose AT-BSN, which can better balance the suppression of noise correlation and the destruction of image's local spatial structure by applying asymmetric blind-spots during training and inference.
    \item We propose a Blind-Spots Based Multi-Teacher Distillation strategy, which significantly improves performance by distilling a lightweight student network from teachers with different blind-spots sampled from the teacher network distribution.
    \item Experimental results on multiple real-world datasets show our method achieves state-of-the-art performance, with clear advantages in computational complexity and preservation of high-frequency texture details.
\end{itemize}

%% file: sec/2_related.tex
\section{Related Works}
\label{sec:related}

\paragraph{Supervised Image Denoising.}
Deep learning has made remarkable advances in image denoising in recent years. Zhang \etal ~\cite{DnCNN} introduced DnCNN, the first CNN-based method for supervised denoising, which significantly outperformed traditional methods~\cite{Chambolle2004AnAF,BM3D,Elad2006ImageDV,WNNM,Vese2003ModelingTW}.The following work aimed to enhance the performance of supervised denoising, such as FFDNet~\cite{FFDNet}, CBDNet~\cite{CBDNet}, RIDNet~\cite{RIDNet}, DANet~\cite{DANet}, FADNet~\cite{FADNet}, and so on. However, supervised-based methods require large amounts of aligned clean-noisy pairs as training data, which are usually difficult and costly to obtain in formal scenarios.

\noindent
\textbf{Unpaired Image Denoising.}
To tackle the challenge in supervised learning, some generative-based~\cite{gans} approaches synthesize noisy samples from clean images~\cite{GAN2GAN,GCBD,UIDNet,C2N,DBSN,fu2023srgb}. The simulated clean-noisy pairs can be further used to train a supervised denoising model. However, the performance of unpaired image denoising methods can be limited when the existing clean images do not match the distribution of the current scene.

\noindent
\textbf{Self-Supervised Image Denoising.}
Lehtinen \etal ~\cite{Noise2Noise} proposed Noise2Noise, which demonstrated that a denoising network could be trained with two independent noisy observations of the same scene. However, even if Noise2Noise relaxes the clean image requirement, obtaining two aligned noisy images in real-world scenarios remains difficult. Noise2Void~\cite{Noise2Void} and Noise2Self~\cite{Noise2Self} proposed a blind-spot strategy to learn denoising from only single noisy images. Further works~\cite{Laine2019,DBSN} extended the paradigm to blind-spot network (BSN) through shifted convolutions~\cite{Laine2019} and dilated
convolutions~\cite{DBSN}. Blind-spot means the network is designed to denoise each pixel from its surrounding spatial neighborhood without itself, thus, the identity mapping to the noisy image itself can be avoided. Noisier2Noise~\cite{Noisier2Noise}, Noisy-As-Clean (NAC)~\cite{NAC}, Recorrupted-to-Recorrupted (R2R)~\cite{R2R}, and IDR~\cite{IDR} generated noisy training pairs by adding synthetic noise to given noisy inputs.
Recently, Neighbor2Neighbor~\cite{Neighbor2Neighbor} proposed to subsample the noisy input images to obtain noisy pairs for Noise2Noise-like training. Blind2Unblind~\cite{Blind2Unblind} proposes a global-aware mask mapper and re-visible loss to fully excavate the information in the blind-spot for Noise2Void-like training.

\begin{figure*}[t]
    \centering
    \includegraphics[width=1\linewidth]{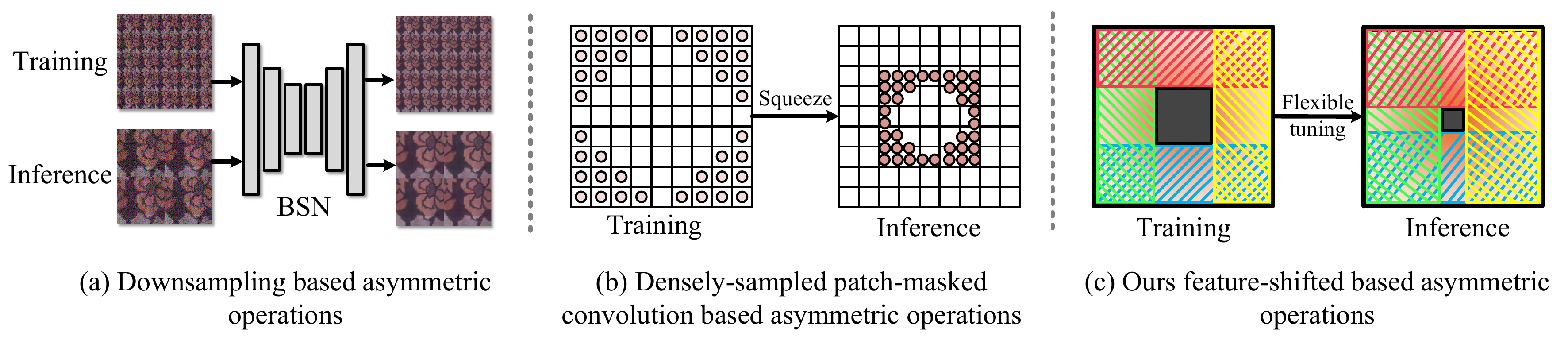}
    \caption{Three kinds of asymmetric operations during training and inference. 
    Our scheme can flexibly tune the blind-spot size to meet the requirements of training and inference, achieving a balance between noise correlation suppression and local spatial destruction.
}
    \label{fig:three_asym}
\end{figure*}

\noindent
\textbf{Real-World Image Denoising.}
Some works~\cite{SIDD,NIND} attempt to capture clean-noisy pairs in real-world scenarios. Abdelhamed \etal ~\cite{SIDD} carefully took and aligned clean-noisy pairs from different scenes and lighting conditions using five representative smartphone cameras, and proposed the SIDD dataset. These datasets enable supervised methods~\cite{NBNet,P3AN,AINDNet,InvDN,VDN,DANet} to train on real-world clean-noisy pairs. However, constructing real datasets requires tremendous human effort and time. Moreover, real-world noise tends to exhibit spatial correlation, which contradicts the premise of Noise2Noise~\cite{Noise2Noise} that noise follows an independent and identically distributed pattern, rendering it and its subsequent variants unsuitable for direct application to real-world scenarios. In order to apply self-supervised learning to real-world settings, Neshatavar \etal ~\cite{CVFSID} introduced a cyclic multi-variate function to disentangle clean images, signal-dependent noise, and signal-independent noise from noisy images.
However, the method relies on a simple network without residual connections to avoid learning an identity mapping to the noise signal. Additionally, the simple assumption about real-world noise signals has resulted in its vague denoising results. 
Lee \etal ~\cite{APBSN} employed pixel-shuffle downsampling (PD)~\cite{whenAWGN} to disrupt the spatial correlation of noise and introduced different PD stride factors
for training and inference for better performance.
Li \etal ~\cite{li2023spatially} proposed to use a larger blind-neighborhood to suppress the spatial correlation of noise and present a network to extract the texture within the blind-neighborhood region. However, the method still uses a large blind-spot during testing, which requires a lot of training to extract effective information from a distance to reconstruct the central pixel. LG-BPN ~\cite{wang2023lg} proposes to mask the central area of a large convolution kernel to suppress the spatial correlation of noise and proposes a dilated Transformer block to extract global information. However, the introduction of large kernels will bring greater computational overhead. In addition, some methods attempt to improve AP-BSN. Pan \etal \cite{pan2023random} propose random sub-sampling as data augmentation. Jang \etal \cite{jang2023self} utilize information from the blind-spot position by proposing conditional masked convolution. Nevertheless, these downsampling-based methods lack texture details.

%% file: sec/3_method.tex
\section{Methods}
\subsection{Revisit of Various Methods to Disrupt Noise Spatial Correlation}
\label{sec:3-1}

\begin{figure}[t]
    \centering
    \includegraphics[width=1\linewidth]{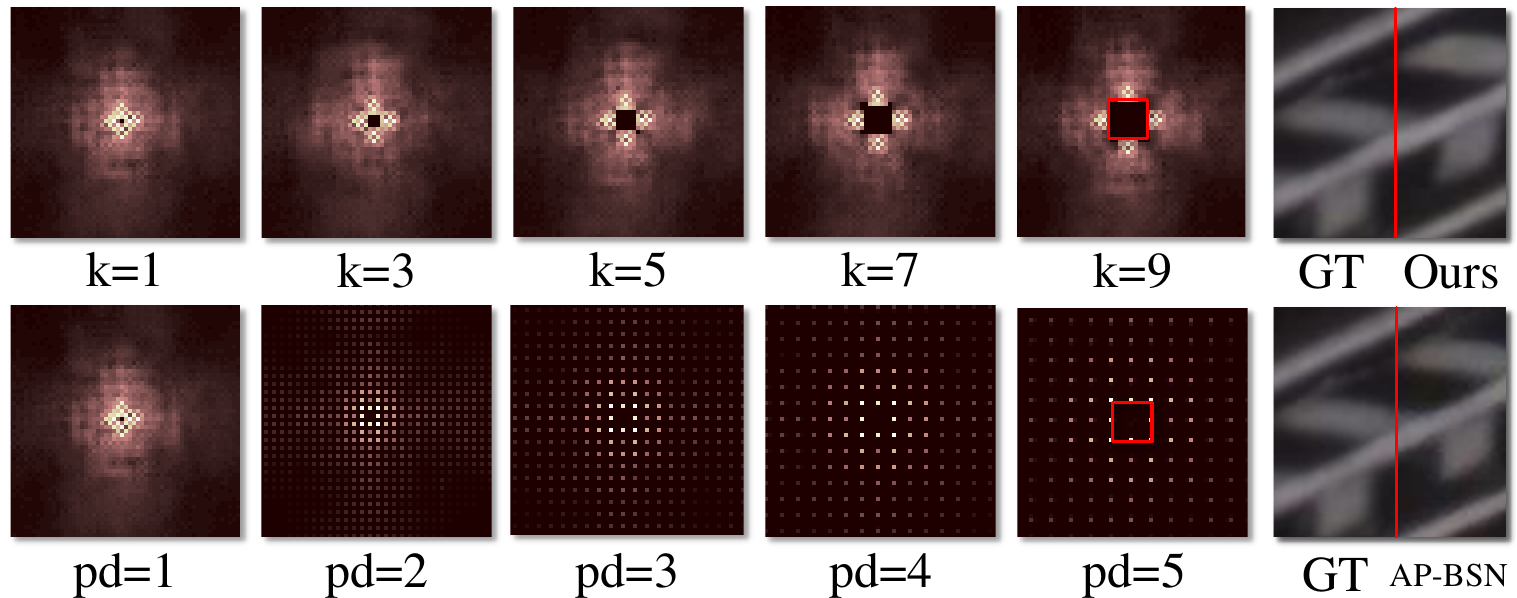}
    \caption{Effective Receptive Field analysis of AP-BSN and our AT-BSN. Each column has the same center blind-spot size. The downsampling operation of AP-BSN can cause aliasing effects.}
    \label{fig:erf}
\end{figure}

Due to the effect of image signal processors (ISP), e.g. image demosaicking~\cite{chatterjee2011noise,jin2020review,park2009case}, real-world noise is generally known to be spatially correlated and pixel-wise dependent. Lee \etal ~\cite{APBSN} analyzed the spatial correlation of real-world noise and found that different camera devices in the SIDD dataset show similar noise behaviors in terms of spatial correlation. According to Lee \etal ~\cite{APBSN}'s analysis, the correlation of noise presents a Gaussian distribution that decays as distance increases. This correlation of noise violates the pixel-wise independent noise assumption of the BSN, rendering it inadequate for real noise removal.

Recently, some methods have been proposed to break the spatial correlation of noise, so that BSN can be used for real-world noise removal. Basically, these methods can be divided into downsampling based approaches and neighborhood masking based approaches.

\noindent
\textbf{Downsampling Based Approaches.}
AP-BSN~\cite{APBSN} first introduced the pixel-shuffle downsampling operation~\cite{whenAWGN} into the self-supervised denoising task to break the noise correlation, and proposed asymmetric PD to balance between noise correlation removal and image structure damage. Jang \etal ~\cite{jang2023self} design a conditional blind-spot network, which selectively controls the blindness of the network to use the center pixel information. By retaining some information of the blind-spot part at test time, this method achieved better results. Further, Pan \etal ~\cite{pan2023random} propose random sub-sampling to address the data-hungry issue of training BSN with real noisy images, and further propose to use sampling difference as a perturbation to improve performance. Regardless of the exact downsampling method used, these methods aim to optimize the BSN $B_{\theta}{(\cdot)}$ mainly by minimizing the following loss functions:
\begin{align}
    \label{Downsamplingloss}
    \mathcal{L}_{down} &= {\Vert D_m^{-1}(B_{\theta}(D_m{(I_{noisy})})) - I_{noisy} \Vert}_1, \nonumber \\
    &or\ {\Vert B_{\theta}(D_m{(I_{noisy})}) - D_m{(I_{noisy})} \Vert}_1,
\end{align}
where $D_m{(\cdot)}$ denotes a certain downsampling method with a factor of $m$, $D_m^{-1}{(\cdot)}$ denotes its inverse operation.
Although the downsampling based methods can effectively break the spatial correlation of noise, the results of these methods are often blurry and lack texture details. According to the Nyquist-Shannon sampling theorem, the fidelity of the results is positively correlated with the sampling density. These methods train on degraded low-resolution images, and their receptive fields are present as sparse and diffuse grids, which makes it difficult to learn structural information. Moreover, these methods still test on low-resolution sub-images, which greatly reduces their sampling density and produces aliasing effects, leading to high-frequency detail loss. Additional time-consuming post-processing is also needed to eliminate aliasing effects.

\noindent
\textbf{Neighborhood Masking Based Approaches.}
The neighborhood masking based schemes attempt to train the denoising network on the original resolution structure, the optimization objective is as follows:
\begin{align}
    \label{fullresloss}
    \mathcal{L}_{neighbor} = {\Vert B^{'}_{\theta}(I_{noisy}) - I_{noisy} \Vert}_1,
\end{align}
where $B^{'}_{\theta}(\cdot)$ denotes a carefully modified BSN, in which the blind-spot size is enlarged.
LG-BPN~\cite{wang2023lg} proposed a densely-sampled patch-masked convolution, which breaks the spatial correlation of noise by masking the center part of a large convolution kernel. LG-BPN also proposed to squeeze the convolution kernel weights during inference to reduce the masked part, to balance between local structure damage and noise correlation removal. Note that this idea is similar to the asymmetric PD in AP-BSN, but the introduction of large convolution kernels brings high computational overhead, and the squeeze of convolution kernel weights is limited to specific convolution kernel sizes, which is inflexible to adjust the masked region. Li\etal ~\cite{li2023spatially} proposed to use a larger blind-neighborhood to break the noise correlation, and trained another network with a receptive field limited to the blind-neighborhood to fill in the information loss within the large blind-neighborhood position. However, this scheme utilizes the same large blind-neighborhood during training and inference, which lacks the consideration of local spatial structure damage.

\begin{figure*}[t]
    \centering
    \includegraphics[width=1\linewidth]{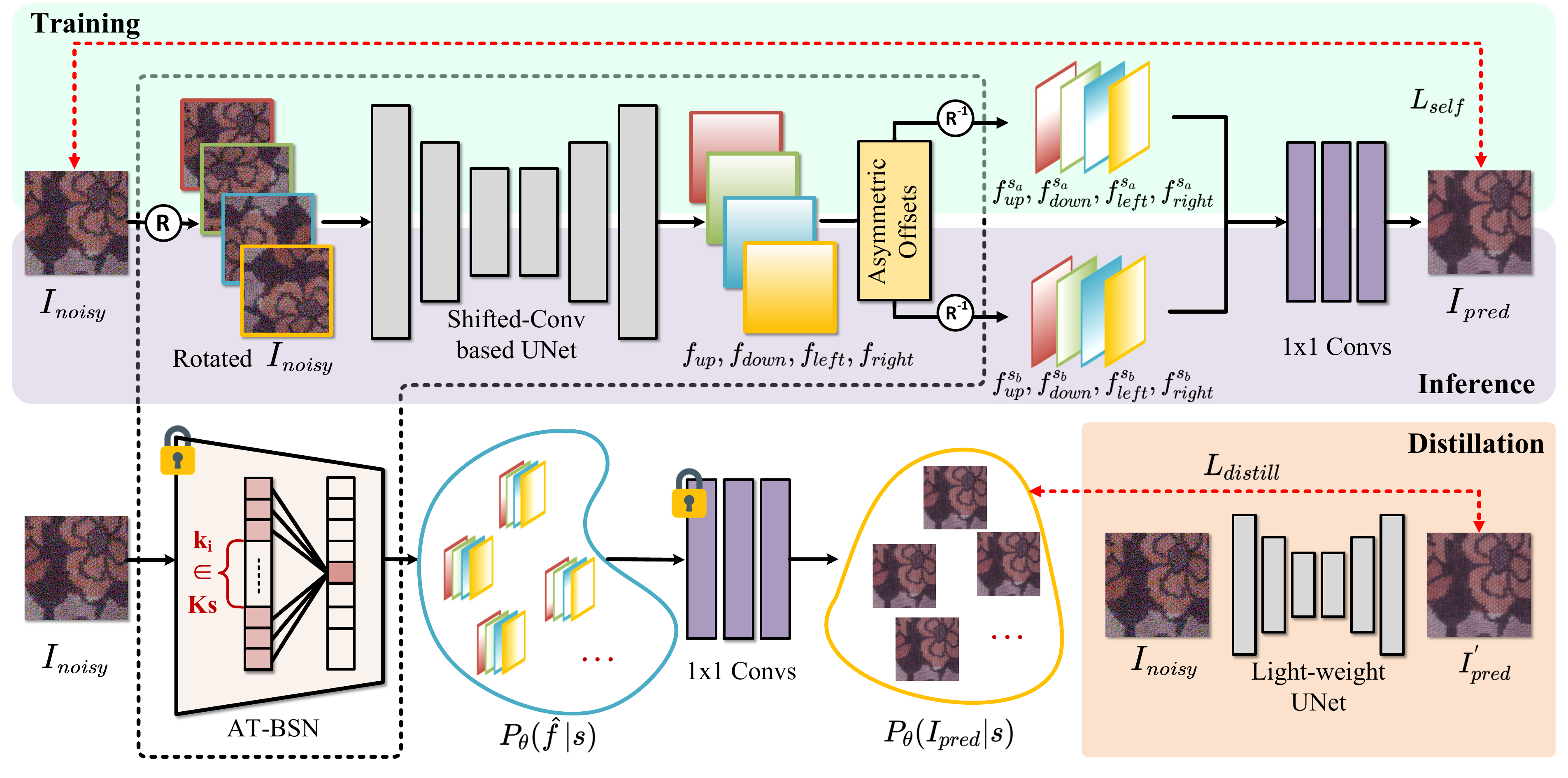}
    \caption{Overview of the proposed AT-BSN framework. We employ asymmetric blind-spots for training and inference to balance the suppression of noise spatial correlation and local information preservation. We regard the trained AT-BSN as a meta-teacher network, generate multiple teacher networks by sampling different blind-spots, and execute multi-teacher distillation on a lightweight network.}
    \vspace{-2mm}
    \label{fig:arch}
\end{figure*}

\noindent
\textbf{Effective Receptive Field Analysis.}
We further compare the effective receptive fields (ERF) of the two schemes to demonstrate the importance of training at full resolution structure.
Noise correlation is generally confined to local regions, according to Lee's statistics~\cite{APBSN}. Therefore, it is unnecessary to disrupt regions beyond the local neighbors. We consider the ERF of the central pixel, and take our method and PD operation in AP-BSN as an example, as shown in Fig.~\ref{fig:erf}. The ERF of AP-BSN is calculated on the subgraph and remapped back to the original resolution structure. 
One can find that the ERF of AP-BSN in the original image manifests as a sparse grid-like pattern, akin to the effect of simple stacking of dilated convolutions~\cite{yu2015multi}. This characteristic comes with similar drawbacks to the dilated convolutions~\cite{chen2017rethinking,wang2018understanding,yu2015multi}, namely \textbf{1)} the loss of local information, posing challenges for the model to learn clues from the grid-like discontinuous sub-image for recovering the clean signal, and \textbf{2)} long-ranged information might be not relevant. Moreover, the loss of information continuity outside the center blind-spot can also introduce aliasing artifacts.
Nevertheless, it is apparent that our method only loses a portion of the information within the blind-spot area, while the ERF outside the blind-spot remains unaffected. This inspires us to set the size of blind-spot to 9 during training.
Due to the higher correlation of the signal compared to the noise, the central pixel can be recovered using the pixels outside the blind-spot that are less correlated with it in the noise domain. It is worth noting that the size of the blind-spot can be minimized during inference to reduce information loss.

Based on the analysis of the existing approaches, we draw the following conclusions. In order to recover clean images with clear textures from noisy images, \textbf{the following two points} are crucial:
\begin{itemize}
    \item[{1)}] According to the Nyquist-Shannon sampling theorem, both the training and inference stages of the network need to be conducted on original input resolution structure to ensure sampling density.
\item[{2)}] Asymmetric operation during training and inference is crucial to strike a balance between the disruption of noise spatial correlation and the destruction of local spatial structure. 
\end{itemize}
The quantitative analysis of the second point can be found in the supplementary materials.
Based on these two observations, we propose AT-BSN, a blind-spot network that can flexibly adjust the blind-spot size during training and inference. Fig.~\ref{fig:three_asym} shows three schematic diagrams of asymmetric operations during training and inference. Compared with Fig.~\ref{fig:three_asym} (a), AT-BSN operates on original resolution structure to maximize sampling density. Compared with Figure Fig.~\ref{fig:three_asym} (b), AT-BSN has a lower computational cost and can adjust the blind-spot size at will, which prompts us to further propose a multi-teacher knowledge distillation strategy based on various blind-spots to further improve performance and reduce network complexity.

\subsection{Tunable Blind-Spot}
BSN~\cite{Noise2Void,APBSN,DBSN} is designed to denoise each pixel from its surrounding spatial neighborhood without itself. 
Typically, BSN can be constructed through shifted convolutions~\cite{Laine2019} or dilated convolution~\cite{DBSN}. 

\noindent
\textbf{Restricted Receptive Fields.}
Our AT-BSN is inspired by Laine's approach~\cite{Laine2019}, which combines four branches with restricted receptive fields, each of which is limited to a half-plane that excludes the central pixel.
For a $h \times h$ convolution kernel, we append $d=\lfloor h/2 \rfloor$ rows of zeros at the top of the feature map, apply the convolution, and finally crop the last $d$ rows of the feature map.
For the $2\times2$ pooling layer, we pad the top of the feature map and crop the last row of it before pooling.

\begin{figure}[t]
    \centering
    \includegraphics[width=0.98\linewidth]{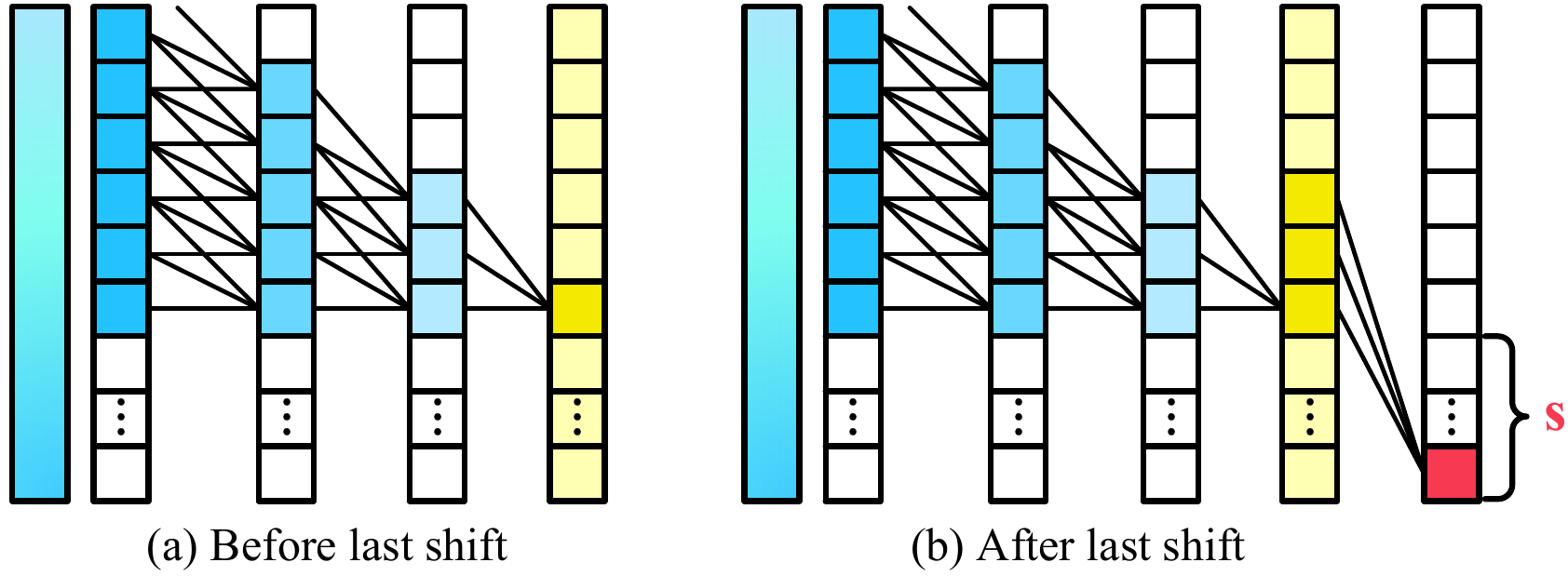}
    \caption{Implementation principle of the tunable blind-spot. 
    }
    \vspace{-3.5mm}
    \label{fig:shift}
\end{figure}

\noindent
\textbf{Tunable Blind-Spot.}
After applying the shifted-conv based UNet to $I_{noisy}$, we obtain the resulting feature map denoted as $f_{up}$, whose receptive field is fully contained within an upward half-plane, including the center row. See Fig.~\ref{fig:shift} for more details.

We further shift the feature map $f_{up}$ downward by $s$ pixels, resulting in a shifted feature map $f_{up}^s$. 
\begin{align}
    \label{shift}
    f_{up}^s = M(f_{up}; s),
\end{align}
where $M(\cdot; s)$ denotes the shift operation of a offset $s$.
At this point, the receptive field of the central pixel only includes the positions beyond $s$ rows above the current location. 
Moreover, \textbf{the use of $M(\cdot; s)$ on the feature domain is decoupled from feature extraction, which grants it more flexibility}.
To expand the receptive field of a pixel to all directions around it, we rotate the input image by multiples of $90^{\circ}$ and feed them into the network. This results in feature maps $f_{up}^s$, $f_{down}^s$, $f_{left}^s$, and $f_{right}^s$.
Finally, the four feature maps are rotated to the correct orientation and linearly combined through several $1\times1$ convolutions to produce the final output $I_{pred}$.
\begin{align}
    \label{conv1x1}
    I_{pred} = Conv_{1\times1}([\hat{f}_{up}^s, \hat{f}_{down}^s, \hat{f}_{left}^s, \hat{f}_{right}^s]),
\end{align}
where $[,]$ denotes feature concatenation, $\hat{f}^s$ denotes the corresponding $f^s$ in the correct orientation.

Now a blind-spot area with a length of $k=2s-1$ is established in the center around each pixel. We can freely tune the size $k$ of the blind-spot by adjusting the shift factor $s$ of the feature map $f^s$ before the $1\times1$ convolutions. So far, we have achieved a BSN with a tunable blind-spot size. We denote the main network parameterized by $\theta$ as $F_{\theta}(\cdot; s)$, the entire process can be formulated as:
\begin{align}
    \label{wholebsn}
    I_{pred} = Conv_{1\times1}{(F_{\theta}(I_{noisy}; s))}.
\end{align}

\subsection{Asymmetric Blind-Spots}

Based on the fact that the noise correlation is less than the signal correlation, we can use appropriate blind-spot $k$ to suppress the noise correlation while minimizing the impact on signal correlation. The central pixel within a large blind-spot can be inferred from pixels outside the blind-spot that are less correlated with it in the noise domain. 
We minimize the following loss to train the network:
\begin{align}
    \label{atbsn-self}
    \mathcal{L}_{self} &= {\Vert Conv_{1\times1}{(F_{\theta}(I_{noisy}; s))} - I_{noisy} \Vert}_1 \nonumber \\
    &= {\Vert I_{pred} - I_{noisy} \Vert}_1.
\end{align}
Following previous works, we use $L^1$ norm for better generalization \cite{APBSN}. In practice, we choose $k=9$, that is, $s=4$, during training. 

We propose to achieve a balance between training and inference by employing asymmetric blind-spots, so that the trade-off between noise correlation removal and image structure damage can be achieved.
Since larger blind-spots have already been utilized during training to enable the BSN to learn to denoise, we can select smaller blind-spots during inference to minimize information loss. We denote the $k$ used in training and inference as $k_a$ and $k_b$, respectively. Similar notation rule is also used for $s$.
In Sec.~\ref{sec_abla}, we will demonstrate the robustness of our approach to different blind-spot combinations.

\begin{table*}[t]
\setlength\tabcolsep{2.6pt}
\begin{center}
\small
    \begin{tabular}{lcccccccc}
    \toprule
    &\multirow{2}{*}{Methods}&  
    \multicolumn{2}{c}{SIDD Benchmark}&\multicolumn{2}{c}{SIDD Validation }&\multicolumn{2}{c}{DND Benchmark} \\
    \cmidrule(lr){3-4} \cmidrule(lr){5-6} \cmidrule(lr){7-8} & & PSNR$^\uparrow$ (dB)  & SSIM$^\uparrow$ & PSNR$^\uparrow$ (dB) & SSIM$^\uparrow$ & PSNR$^\uparrow$ (dB) & SSIM$^\uparrow$ \\
    \midrule
    \multirow{2}{*}{Non-Learning}& BM3D\cite{BM3D}   & 25.65& 0.685 & 31.75 & 0.706 & 34.51 & 0.851      \\
    & WNNM\cite{WNNM}   & 25.78 & 0.809 & 26.31 & 0.524 & 34.67 & 0.865      \\
    \midrule
    \multirow{3}{*}{Supervised}& DnCNN\cite{DnCNN}  & 37.61& 0.941& 37.73 & 0.943 & 37.90 & 0.943      \\
    & CBDNet\cite{CBDNet}   & 33.28 & 0.868 & 30.83 & 0.754 & 38.05 & 0.942      \\
    & RIDNet\cite{RIDNet}   & 37.87 & 0.943& 38.76 & 0.913 & 39.25 & 0.952      \\
    & VDN\cite{VDN}   & 39.26& 0.955 & 39.29 & 0.911 & 39.38 & 0.952      \\
    & AINDNet(R)\cite{AINDNet}   & 38.84& 0.951& 38.81 & - & 39.34 & 0.952      \\
    & DANet\cite{DANet}  &39.25 &0.955 & 39.47 & 0.918 & 39.58 & 0.955      \\
    & InvDN\cite{InvDN}  &39.28 & 0.955& 38.88 & - & 39.57 & 0.952      \\
    \midrule
    \multirow{2}{*}{Unpaired}
    & GCBD\cite{GCBD}   &- &- & - & - & 35.58 & 0.922      \\
    & D-BSN\cite{DBSN} $+$ MWCNN\cite{MWCNN}   & -& -& - & - & 37.93 & 0.937      \\
    & C2N\cite{C2N}   & 35.35& 0.937& 35.36 & 0.932 & 37.28 & 0.924      \\
    \midrule
    \multirow{3}{*}{Self-Supervised} & Noise2Void\cite{Noise2Void}   & 27.68&0.668 & 29.35 & 0.651 & - & -      \\
    & Laine-BSN\cite{Laine2019}   &- &- & 23.80$^\diamond$ & 0.493$^\diamond$ & - & -      \\
    & Noise2Self\cite{Noise2Self}   & 29.56&0.808 & 30.72 & 0.787 & - & -      \\
    & NAC\cite{NAC}   & -& -& - & - & 36.20 & 0.925      \\ 
    & R2R\cite{R2R}   & 34.78&0.898 & 35.04 & 0.844 & - & -      \\
    & CVF-SID\cite{CVFSID}   & 34.43 / 34.71$^\dagger$& 0.912 / 0.917$^\dagger$& 34.51 & {\color{blue}\textbf{0.941}} & 36.31 / 36.50$^\dagger$ & 0.923 / 0.924$^\dagger$      \\
    & AP-BSN $+$ R$^3$\cite{APBSN}   & 35.97 / 36.91$^\dagger$&0.925 / 0.931$^\dagger$ & 35.76 & - & \phantom{xx}-\phantom{xx} / 38.09$^\dagger$ & \phantom{xx}-\phantom{xx} / 0.937$^\dagger$      \\
    & C-BSN\cite{jang2023self}   & 36.82 / \phantom{xx}-\phantom{xxx}&0.934 / \phantom{xx}-\phantom{xxx} &  36.22& 0.935 & {\color{red}\textbf{38.45}} / {\color{blue}\textbf{38.60}}$^\dagger$ & {\color{red}\textbf{0.939}} / {\color{blue}\textbf{0.941}}$^\dagger$      \\
    & SDAP (E)\cite{pan2023random}   & 37.24 / {\color{blue}\textbf{37.53}}$^\dagger$&{\color{blue}\textbf{0.936}} / {\color{blue}\textbf{0.936}}$^\dagger$ &  37.30& 0.894 & 37.86 / 38.56$^\dagger$ & 0.937 / 0.940$^\dagger$      \\
    & LG-BPN\cite{wang2023lg}   & 37.28 / \phantom{xx}-\phantom{xxx}&{\color{blue}\textbf{0.936}} / \phantom{xx}-\phantom{xxx} &  37.32& 0.886 & \phantom{xx}-\phantom{xx} / 38.43$^\dagger$ & \phantom{xx}-\phantom{xx} / {\color{red}\textbf{0.942}}$^\dagger$      \\
    & Spatially-Adaptive (UNet)\cite{li2023spatially}   & {\color{blue}\textbf{37.41}} / 37.37$^\dagger$&0.934 / 0.929$^\dagger$ &  {\color{blue}\textbf{37.39}}& 0.934 & 38.18 / 38.58$^\dagger$ & {\color{blue}\textbf{0.938}} / 0.936$^\dagger$      \\ 
    & \textbf{AT-BSN (Ours)}  &36.73 / 36.74$^\dagger$ & 0.924 / 0.925$^\dagger$  & 36.80  & 0.934  &  37.76 / 38.19$^\dagger$ &  0.934 / 0.939$^\dagger$    \\
    & \textbf{AT-BSN (D) (Ours)}  & {\color{red}\textbf{37.77}} / {\color{red}\textbf{37.78}}$^\dagger$ & {\color{red}\textbf{0.942}} / {\color{red}\textbf{0.944}}$^\dagger$ & {\color{red}\textbf{37.88}} & {\color{red}\textbf{0.946}} & {\color{blue}\textbf{38.29}} / {\color{red}\textbf{38.68}}$^\dagger$ & {\color{red}\textbf{0.939}} / {\color{red}\textbf{0.942}}$^\dagger$      \\
    \bottomrule
    \end{tabular}
\end{center}
\vspace{-3.5mm}
\caption{Comparison among different denoising methods on real-world datasets. We report the official results from the benchmark website or related paper. The $\dagger$ marks indicate the method is trained directly on the corresponding benchmark dataset in a fully self-supervised manner. The $\diamond$ marks indicate the result is measured by ourselves. }
\vspace{-1mm}
\label{tab:dabiao}
\end{table*}

\subsection{Blind-Spots Based Multi-Teacher Distillation}
\label{sec:3p4}
While larger blind-spots can more effectively suppress spatial correlations between neighboring
noise signals, they also result in more loss of information. Conversely, smaller blind-spots exhibit an opposite trend.

To better integrate the advantages of our tunable blind-spot nature, we propose a blind-spot based multi-teacher distillation strategy.
Under different blind-spot sizes (or different $M(\cdot; s_i)$), the features extracted by AT-BSN satisfy the distribution $P_{\theta}(\hat{f}|s)$, and the restored clear images follow $P_{\theta}(I_{pred}|s)$. We consider the trained AT-BSN as a meta-teacher network that can generate many potential teacher networks almost \textbf{cost-free} by adjusting the size of blind-spots. Therefore, we obtain multiple potential teacher networks, where different teachers can provide different knowledge, i.e., different teachers handle smooth/texture areas differently (detailed analysis can be found in the supplementary materials). This characteristic is a key strength of our approach, allowing the student network to learn from various teachers.

Specifically, we pass $I_{noisy}$ through the trained network to get the feature $f$. Subsequently, we sample multiple $k$ from $Ks\in \{0,1,3,...,2S-1\}$ (where $S$ denotes the preset maximum offset), apply $M(\cdot; s_i)$ to $f$ multiple times, and obtain the features $\hat{f}^{s_i}$ under different blind-spot sizes. Then we apply trained $1\times1$ convolutions to get the clear images $I^{s_i}_{pred}$. Finally, we use $I^{s_i}_{pred}$ to distill a lightweight non-blind-spot network $N_{\theta}(\cdot)$. In order for the student network to learn fairly from $P_{\theta}(I_{pred}|s)$, we do not distinguish between different teacher signals explicitly. We set the same weight $\alpha_i=1$ for each teacher.
The student network is distilled by optimizing the following objective:
\begin{align}
    \label{atbsn-distill}
    \mathcal{L}_{distill} = \sum_{s_i\in K_s}\alpha_{i}{\Vert N_{\theta}(I_{noisy}) - sg(I^{s_i}_{pred}) \Vert}_1,
\end{align}
where $sg(\cdot)$ denotes stop gradient operation.
Under the multi-teacher distillation scheme, our student network can be lightweight and avoid the additional computational cost brought by the rotation operation of the BSN. 

Moreover, the distillation itself is also computationally efficient, as multiple teachers share the same feature $f$. The specific complexity analysis can be found in the supplementary materials.
The overall scheme of our methods can be found in Fig.~\ref{fig:arch}.

\begin{figure*}[t]
    \centering
    \includegraphics[width=0.95\linewidth]{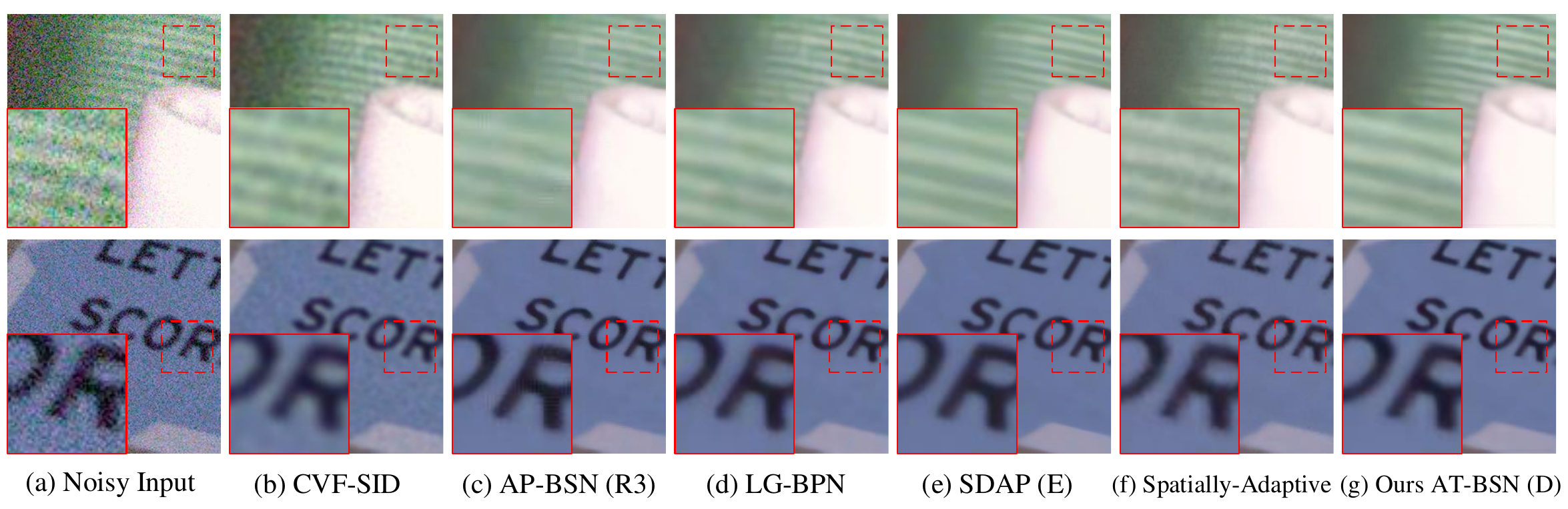}
    \vspace{-2mm}
    \caption{Quantitative comparisons on SIDD validation dataset.}
    \vspace{-3.5mm}
    \label{fig:sidd}
\end{figure*}

%% file: sec/4_experiment.tex
\section{Experiments}
\label{sec:exp}
\subsection{Experimental Configurations}

\noindent
\textbf{Real-World Datasets.}
We conduct experiments on two real-world image denoising datasets, SIDD~\cite{SIDD} and DND~\cite{DND}.
SIDD-Medium training dataset consists of 320 clean-noisy pairs captured under various scenes and illuminations. The SIDD validation dataset contains 1280 noisy patches with a size of $256\times256$ for performance evaluation. 
DND benchmark consists of 50 noisy images captured with consumer-grade cameras of various sensor sizes and does not provide clean images. DND dataset is captured under normal lighting conditions compared to the SIDD dataset, and therefore presenting less noise. We adopt PSNR and SSIM metrics to evaluate our method. We set $k_a=9$ and $k_b=3$ for training and inference, respectively. For multi teacher distillation, we sample blind-spots from $Ks\in{ \{0,1,3,5,7,9,11\} }$. More implementation details can be found in supplementary materials.

\subsection{Comparisons for Real-World Denoising}

\noindent
\textbf{Quantitative Measure.}
Tab.~\ref{tab:dabiao} presents quantitative comparisons with other methods. We denote the distilled network as AT-BSN (D). Note that AT-BSN (D) is actually a Non-BSN despite its name. As a self-supervised algorithm, our method outperforms all existing unpaired and self-supervised methods, achieving state-of-the-art performance. Our results without $\dagger$ marks indicate we employ the model trained on SIDD-Medium directly on the benchmark. These results show the generalization ability of our method. Our results with $\dagger$ show the potential of fully self supervised learning.
Furthermore, the results of AT-BSN (D) demonstrate the potential to enhance performance by integrating the advantages of different blind-spot sizes through multi teacher distillation.

\noindent
\textbf{Qualitative Measure.}
Fig.~\ref{fig:sidd} presents the qualitative comparisons. Our method is capable of preserving the most texture details. In addition, results of downsampling-based methods~\cite{APBSN,pan2023random} tend to transition smoothly. LG-BPN~\cite{wang2023lg}, Spatially Adaptive~\cite{li2023spatially}, and our AT-BSN, as training on the original resolution structure, can preserve better results. Specifically, although AP-BSN uses R$^3$ post-processing, its visualization still exhibits aliasing effects. More SIDD and DND benchmark results are in the supplementary materials.

\begin{table}[t] 
\setlength\tabcolsep{4.6pt}
\small
\begin{center}
    \begin{tabular}{lccccc}
    \toprule
    \multirow{2}{*}{Blind-Spots}& \{1,3,   & \{0,1,3,  & \{0,1,3, & \{0,1,3& \{0,1,3,5\\
    & 5,7\}   & 5,7\}  & 5,7,9\} & 5,7,9,11\}& 7,9,11,13\}      \\
    \midrule
    PSNR    & 37.31   & 37.32  & 37.41 & \textbf{37.47} &37.40       \\
    SSIM    & 0.943   & 0.943  & 0.944  & \textbf{0.945} &0.944      \\
    \bottomrule
    \end{tabular}
\end{center}
\vspace{-5mm}
\caption{Ablation study of the ensemble of teacher networks.}
\label{tab:ManyBlindSpots}
\end{table}

\begin{table}[t] 
\small
\vspace{-1.5mm}
\begin{center}
    \begin{tabular}{lccc}
    \toprule
    \multirow{2}{*}{}& Mean Teacher   & Multi Teacher  & Param  \\
    \midrule
    Student A    & 36.79 / 0.942   & 36.98 / 0.943  &  \textbf{0.12 M}       \\
    Student B    & 37.60 / 0.945   & 37.76 / 0.946  & 0.86 M        \\
    Student C    & \textbf{37.72 / 0.945}   & \textbf{37.88 / 0.946}  & 1.02 M       \\
    \bottomrule
    \end{tabular}
\end{center}
\vspace{-5mm}
\caption{Ablation study of different distillation methods on students with different parameters.}
\label{tab:mean_and_multi}
\end{table}

\begin{table}[t] 
\setlength\tabcolsep{2.6pt}
\small
\vspace{-1.5mm}
\begin{center}
    \begin{tabular}{lccc}
    \toprule
    \multirow{2}{*}{Methods}& Params$^\downarrow$   & MACs$^\downarrow$  & PSNR$^\uparrow$  \\
    & (M)  & (G)  & (dB) \\
    \midrule
    CVF-SID
    \cite{CVFSID}    & 1.19   & 311.44    & 34.81      \\
    AP-BSN $+$ R$^3$\cite{APBSN}    & 3.66   & 7653.97    & 36.48      \\
    SDAP (E)\cite{pan2023random}    & 3.66   & 1628.57    & 37.30     \\
    LGBPN\cite{wang2023lg}    & 4.56   & 12168.22    & 37.32     \\
    Spatially-Adaptive (UNet)\cite{li2023spatially}    & 1.08   & 70.11    & 37.39     \\
    \textbf{AT-BSN (Ours)}    & 1.27   & 330.51    & 36.80     \\
    \textbf{AT-BSN (D) (Ours)}    & \textbf{1.02}   & \textbf{48.92}    & \textbf{37.88}     \\
    \bottomrule
    \end{tabular}
\end{center}
\vspace{-5mm}
\caption{Complexity Analysis. The multiplier-accumulator operations (MACs) are measured on $512 \times 512$ patches.}
\label{tab:Complexity}
\vspace{-3mm}
\end{table}

\subsection{Ablation Study}
\label{sec_abla}

\begin{figure}
    \centering
    \includegraphics[width=0.95\linewidth]{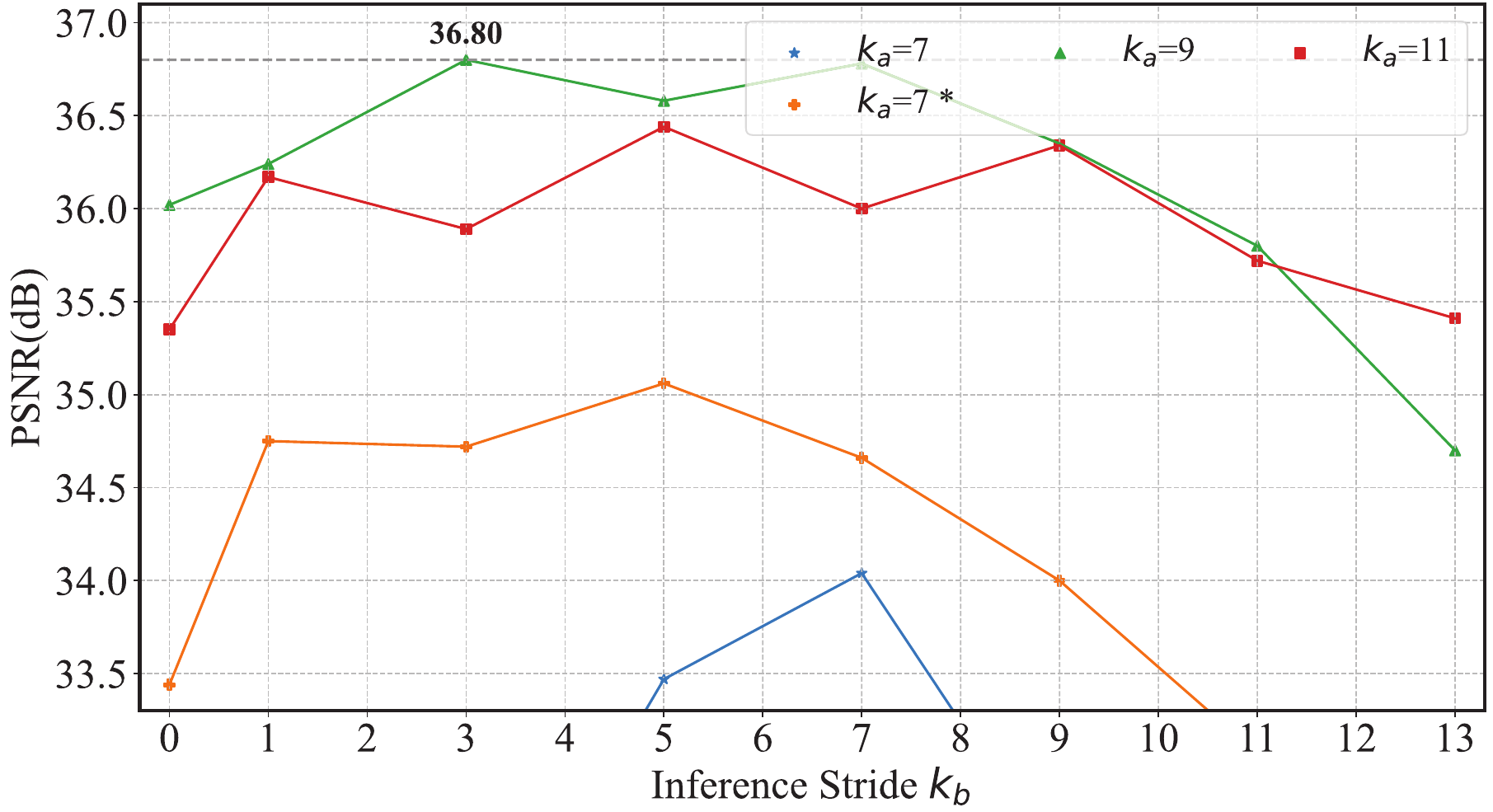}
    \vspace{-3mm}
    \caption{Ablation experiments on the combinations of different blind-spots between training and inference. }
    \vspace{-3mm}
    \label{fig:robust}
\end{figure}

\noindent
\textbf{Analysis of Asymmetric Blind-Spots.}
We perform experiments on the combinations of training blind-spot sizes $k_a\in\{7, 9, 11\}$ and inference blind-spot sizes $k_b\in \{0,1,3,5,7,9,11,13\} $, testing on the SIDD validation dataset. 
From Fig.~\ref{fig:robust}, Our method could achieve the best performance at $k_a=9$ and $k_b=3$. This is due to the noise correlation is fully suppressed during training, the network has learned well denoising ability. During testing, only small blind-spot is needed to suppress the areas with the highest noise correlation, while maximizing the preservation of local information.
Performance degradation is observed in the $k_a=7$ setting. To address this, we introduce early stopping in this setting ($k_a=7*$), resulting in relatively better results. This suggests that a blind-spot with $k_a=7$ can partially suppress spatial noise but not entirely. Increasing epochs leads the network to learn from outside the blind-spot to reconstruct central noise.

\noindent
\textbf{Ensemble of Teacher Networks.}
To investigate how the student network benefits from learning various teachers, we conducted additional ablation experiments. A straightforward approach is to ensemble multiple teacher networks with different blind-spots by averaging their outputs for the final denoising result. Tab.~\ref{tab:ManyBlindSpots} displays the performance achieved by averaging results from teacher networks across different sets $Ks$. Performance tends to increase with larger $Ks$, indicating that teacher networks with distinct blind-spots perform differently across image areas, and weighting them effectively enhances performance. However, excessively large $Ks$ can lead to overly smooth average images, resulting in decreased performance.

\noindent
\textbf{Different Approaches of Distillation. }
We compare two methods of distillation: mean teacher distillation, utilizing the average outputs of various teachers for student training, and multi-teacher distillation, our chosen method in this study. Tab.~\ref{tab:mean_and_multi} presents the outcomes of distilling three UNets with varying parameters using these approaches, with the latter consistently yielding superior performance.
We omit a prior for distinguishing image areas, allowing the student to adaptively capture complementary information from different teachers, learning from multiple perspectives and thus enhancing generalization~\cite{fukuda2017efficient}. Conversely, mean teacher distillation employs pixel averaging as a prior, reducing student network robustness and performance.

\noindent
\textbf{Complexity Analysis. }
Tab.~\ref{tab:Complexity} compares the complexity of different methods. Our method attains superior performance with the lowest computational overhead. It's worth noting that both AP-BSN and LG-BPN utilize R$^3$~\cite{APBSN} operations, substantially increasing computational overhead.

%% file: sec/5_conclusion.tex
\vspace{-2mm}
\section{Conclusions}
\vspace{-1.5mm}

In this paper, we first analyze existing self-supervised denoising techniques, highlighting the importance of training at the original resolution structure and using asymmetric operations. We then introduce a new approach using asymmetric blind-spots to balance noise suppression and spatial structure preservation, and present a blind-spots based multi-teacher distillation strategy. Experimental results show that our method achieves state-of-the-art and is superior in computational complexity and visual effect.

\vspace{-2mm}
\section*{Acknowledgment}
This work was supported by the National Natural Science Foundation of China under grant numbers 62176003, 62088102, and 62306015, and by the Beijing Nova Program under grant number 20230484362.

%% file: sec/X_suppl.tex
\clearpage
\setcounter{page}{1}
\maketitlesupplementary

\renewcommand{\thetable}{S\arabic{table}}
\renewcommand{\thefigure}{S\arabic{figure}}
\renewcommand{\theequation}{S\arabic{equation}}
\renewcommand{\thesection}{S\arabic{section}}

\section{Implementation Details}
\paragraph{Training Settings.}
We adapt UNet-like BSN as our network. We set $k_a=9$ and $k_b=3$ for training and inference, respectively. For the BSN training, we follow settings of previous work~\cite{li2023spatially}. The input images are cropped into $256 \times 256$ patches with a batch size of 8 and augmented with random flipping and rotation. The network is optimized with an Adam optimizer with a learning rate of $3\times10^{-4}$ and $[\beta_1, \beta_2]$ of $[0.9, 0.999]$. The BSN is trained for 400k iterations, and the learning rate is decreased to zero with cosine annealing scheduler. For multi teacher distillation, we sample blind-spots from $Ks\in{ \{0,1,3,5,7,9,11\} }$ for SIDD dataset, and $Ks\in{ \{0,1,3,5,7,9\} }$ for DND dataset as the noise intensity in the DND dataset is relatively small. We crop the input images into $128 \times 128$ patches with a batch size of 8. We set the initial learning rate to $3\times10^{-4}$ and decrease it to zero with cosine annealing scheduler during 100k iterations. 

\paragraph{Network Structure.}
We use a non-blind-spot network (NBSN) during multi teacher distillation, which don't use any shift or rotation operations and remove the final $1\times1$ convolutional layers. Specifically, We show the performance of student networks with three different parameters after distillation in main paper Tab.3. The student A is a lightweight network consisting of 2 downsampling and 2 upsampling layers, each containing only 1 convolution layer. The student B is simillar to A, but uses 3 convolution layers in the downsampling layer and 4 convolution layers in the upsampling layer. The student C increase both the upsampling and downsampling layers to 5, and uses 2 convolution layer and 1 convolution layers in them, respectively. We can find that even small networks can significantly benefit from multi teacher distillation. In addition, increasing the number of downsampling layers or convolutional layers to increase network depth can further improve performance, with the latter achieving better performance.

\paragraph{Details of Distillation.}
Our multi-teacher distillation training process exhibits lower overall complexity compared to training process of AT-BSN. As illustrated in Fig.4 and Sec.3.4, only the meta-teacher (a shifted UNet) needs to compute four directional features once, then produce 7 offset features with negligible shift operations. These features, considered as outputs from potential teachers with different blind spots, undergo final shallow ${1\times1 }$ Convs to yield 7 distillation labels. Thus the total distillation cost is $4$*$MAC_{shifted\_unet}$+$7$*$MAC_{1\times1 }$+$MAC_{light\_unet}$, not $4$*$7$*$(MAC_{shifted\_unet}$+$MAC_{1\times1 })$+$MAC_{light\_unet}$. All experiments were done on a RTX 3080. Following the distillation settings above ($batch\_size$=$8$, $patch$=$128$, $total\_iterations$=$100k$), the distillation finished in around 2.5 hours with speed of 11.1 iters/s and 2.4G GPU memory. In contrast, the Shifted UNet, with the same settings, requires around 3.2 hours to finish 100k iterations with speed of 8.3 iters/s and 5.7G GPU memory. Notably, during distillation, no gradient computation is needed for meta-teacher, and features in four directions are unnecessary for student. Furthermore, in practice, we can speed up the distillation process by loading part of the weights from the trained BSN to initialize the student C as they are almost identical, except for the absence of the last $1\times 1$ convolutions and the difference in the output channel on the last upsampling layer. It should be noted that the shift and rotation operations within BSN only affect the feature maps and do not impact the application of BSN weights to NBSN.

\begin{table}[t] 
\setlength\tabcolsep{1.4pt}
\small
\begin{center}
    \begin{tabular}{lccccc}
    \toprule
    \multirow{2}{*}{Methods}& Downsampling   & Patch-masked  &  Feature-shifted \\
    & based   & conv based & based (Ours) &      \\
    \midrule
    Symmetric    & 33.40   & 35.64  & 36.35 ($k_{a/b}=9/9$)     \\
    Asymmetric    & \textbf{34.86}   & \textbf{37.32}  & \textbf{36.80} ($k_{a/b}=9/3$)       \\
    \bottomrule
    \end{tabular}
\end{center}
\caption{Comparison between symmetric and asymmetric designs on SIDD validation dataset.}
\label{tab:asy_sy}
\end{table}

\begin{figure}[t]
    \centering
    \includegraphics[width=0.95\linewidth]{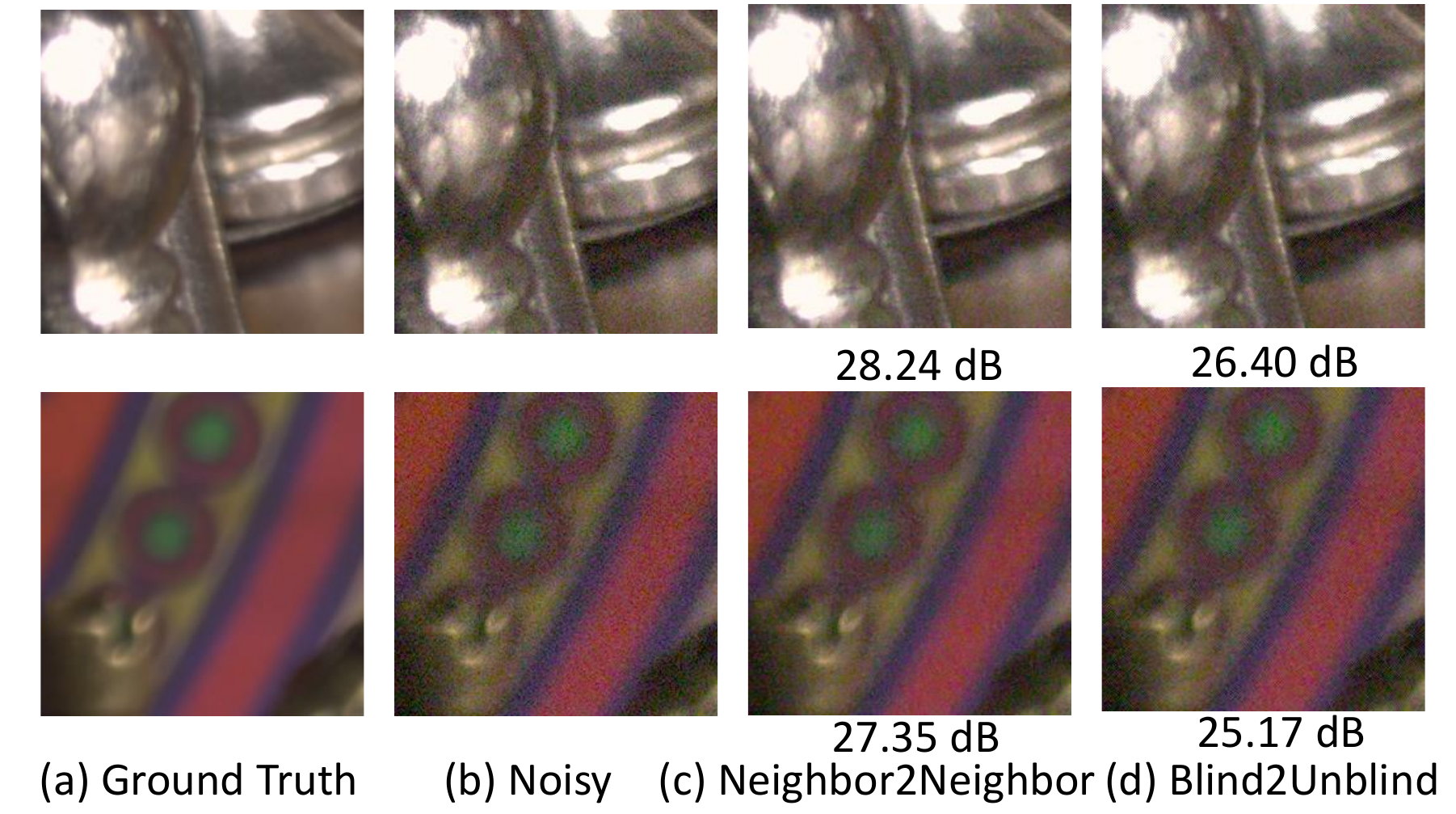}
    \caption{Qualitative comparisons of Neighbor2Neighbor and Blind2Unblind on SIDD validation dataset.}
    \vspace{-5mm}
    \label{fig:supp_nbr2nbr}
\end{figure}
\section{Importance of Asymmetric Design}
Fig.7 compares symmetric and asymmetric BSN, where asymmetric BSN turns to symmetric one with consistent $k$ in training and inference.
Optimal asymmetric BSN performance requires careful use of $k$. 
Tab.\ref{tab:asy_sy} contrasts three asymmetric designs in Fig.2 with their symmetric counterparts (results are from AP-BSN, LG-BPN, and Fig.7), demonstrating the importance of asymmetric design.

\section{Neighbor2Neighbor and Blind2Unblind}
Neighbor2Neighbor and Blind2Unblind are two methods that do not consider the spatial correlation of noise, and in this section we will demonstrate their performance degradation in real-world noise scenarios.

Neighbor2Neighbor~\cite{Neighbor2Neighbor} proposes to subsample the noisy input images to obtain noisy pairs for Noise2Noise-like training. Blind2Unblind~\cite{Blind2Unblind} proposes a global-aware mask mapper and re-visible loss to fully excavate the information in the blind-spot for Noise2Void-like training. Nevertheless, these methods rely on the pixel-wise independent noise assumption~\cite{Noise2Void,Noise2Noise}, which is not satisfied in real-world scenarios. To be specific, the central pixel can be inferred using the neighboring noisy pixels as clues. We retrain both methods on the SIDD-Medium sRGB dataset, and report PSNR values of 25.98 dB and 23.10 dB, respectively. We find that both methods learn an approximate identity mapping that is close to the noisy input itself, as illustrated in Fig.~\ref{fig:supp_nbr2nbr}

\begin{figure}[t]
    \centering
    \includegraphics[width=0.95\linewidth]{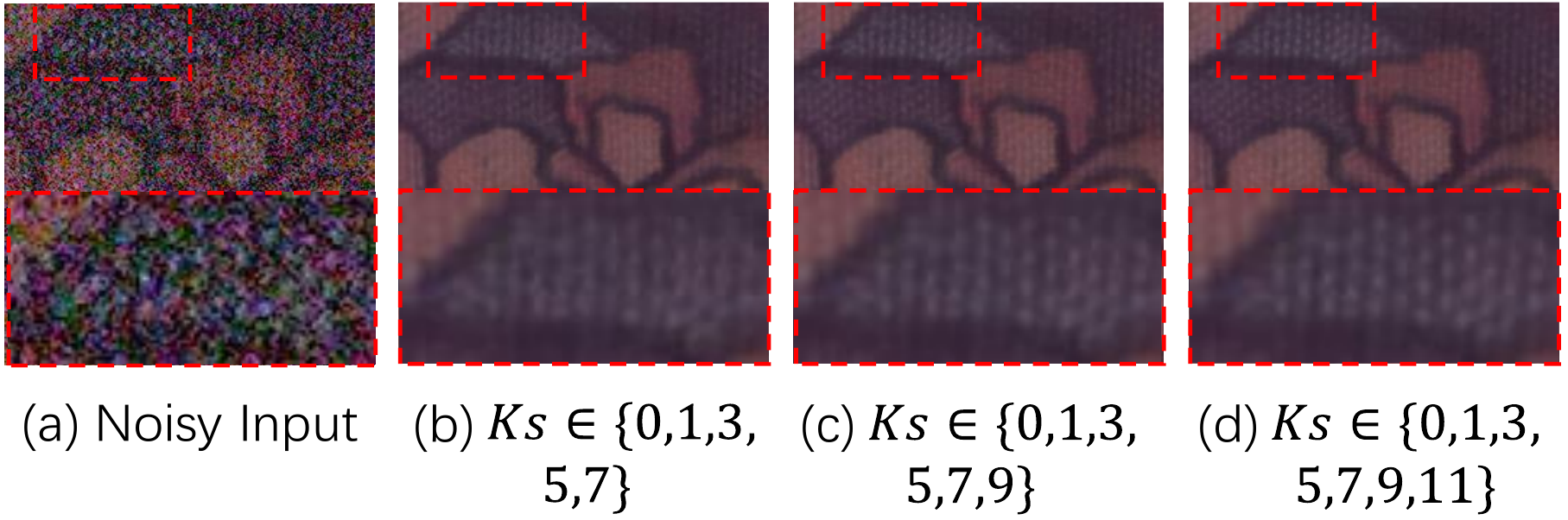}
    \caption{Fusion results of teachers.}
    \label{fig:fusion_results}
\end{figure}
\section{Mean Teacher Distillation and Multi Teacher Distillation}
\begin{figure}[t]
    \centering
    \includegraphics[width=0.92\linewidth]{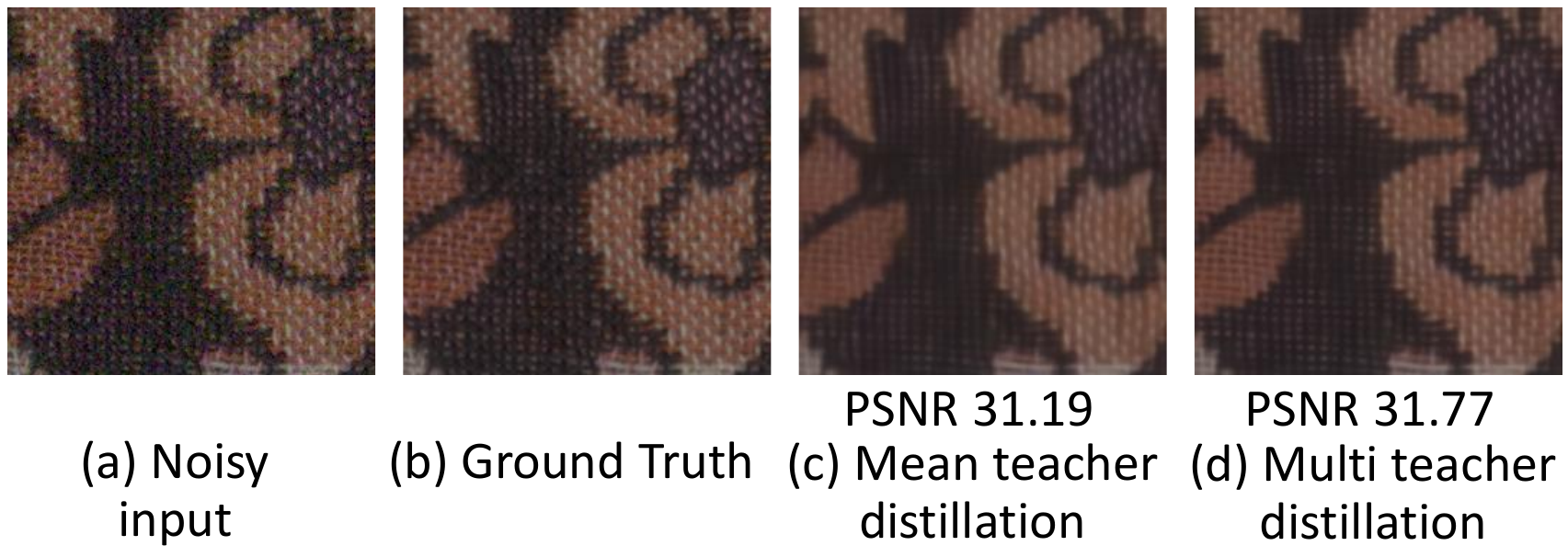}
    \caption{Qualitative comparisons of mean teacher distillation and multi teacher distillation.}
    \label{fig:multimeanfig}
\end{figure}

Fig.~\ref{fig:fusion_results} presents the fusion results of teachers, corresponding to 2-4 columns in Tab.2. One can find gradually improved visual effects as more teachers are fused. Moreover, from Fig.~\ref{fig:multimeanfig}, we can find that result of mean teacher distillation is smoother. Although both mean teacher distillation and multi teacher distillation tend to learn the average image, but we consider all teachers to contribute equally, and the mean teacher cannot capture all the details of teacher distribution, especially when there are significant differences between teachers (e.g. $k_b=1$ and $k_b=11$). On the contrary, we utilize multi teacher distillation to learn different knowledge equally from multiple teacher networks, thereby further improving performance. We avoid explicit region partitioning in distillation loss of \cite{li2023spatially}, ensuring our method not affected by the unstable induction bias of such process, that is, the partitioning heavily relies on initial results. To support this, we imitate the partitioning and utilize outputs of $k$=$9$ $\&$ $k$=$0$ to distill flat and texture area respectively, leading to 37.27 dB, weaker than 37.59 dB with our multi-teacher approach. 

\section{Additional Qualitative Results}
\subsection{Different Combinations of Blind-Spots}
To investigate the impact of different combinations of blind-spot sizes $k$ during training and inference, we conduct additional qualitative experiments. Please note that for the setting of $k_a=7$, we use early stopping to avoid overfitting.

\begin{figure*}[t]
    \centering
    \includegraphics[width=1\linewidth]{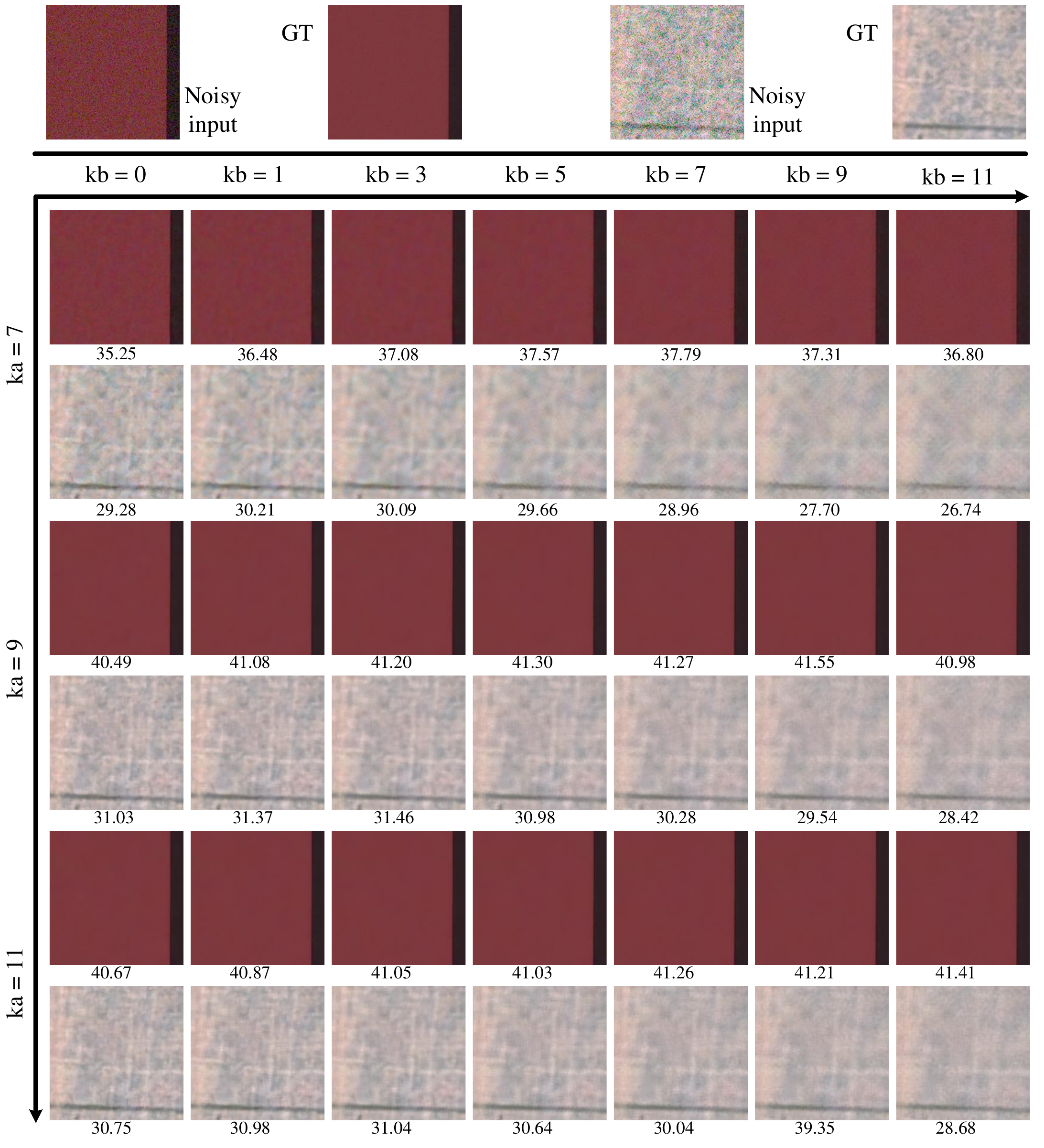}
    \caption{Qualitative results of different combinations of blind-spots during training ($k_a$) and inference ($k_b$) on SIDD validation dataset.}
    \label{fig:combinations}
\end{figure*}

We select a set of images with rich textures and another set with relatively flat textures. From Fig.~\ref{fig:combinations}, We find that under various $k_a$, for flat images, the PSNR of the results is approximately positively correlated with $k_b$. This is because larger blind-spots during testing can effectively suppress noise correlation, and the recovery of flat areas is insensitive to the loss of local information. For images with more textures, the PSNR of the results is approximately negatively correlated with $k_b$. This is because larger blind-spots during testing will lead to the loss of local spatial information, making it difficult to recover local texture details.

Interestingly, we also find that when $k_a$ is larger during training, the destructive effect of larger $k_b$ on texture information becomes smaller. This is because the network trained on larger $k_a$ has stronger ability to find clues from farther places to recover the current pixel, so it can still maintain a certain degree of texture information at larger $k_b$.

Visually, the flat area gradually becomes cleaner with the increase of $k_b$, while the texture area gradually becomes more blurred with the increase of $k_b$. This once again shows that different blind-spots have different denoising effects on flat/texture areas. From the above analysis, we can see that the removal of noise and the preservation of texture details is a dilemma. Our multi-teacher distillation can learn from multiple teacher networks with different blind-spots, thereby achieving a balance in suppressing noise space correlation and maintaining local textures, that is, achieving a balance in denoising flat areas and texture areas, thereby greatly improving performance.

\subsection{More Results on DND and SIDD Datasets}
\begin{figure*}[t]
    \centering
    \includegraphics[width=1\linewidth]{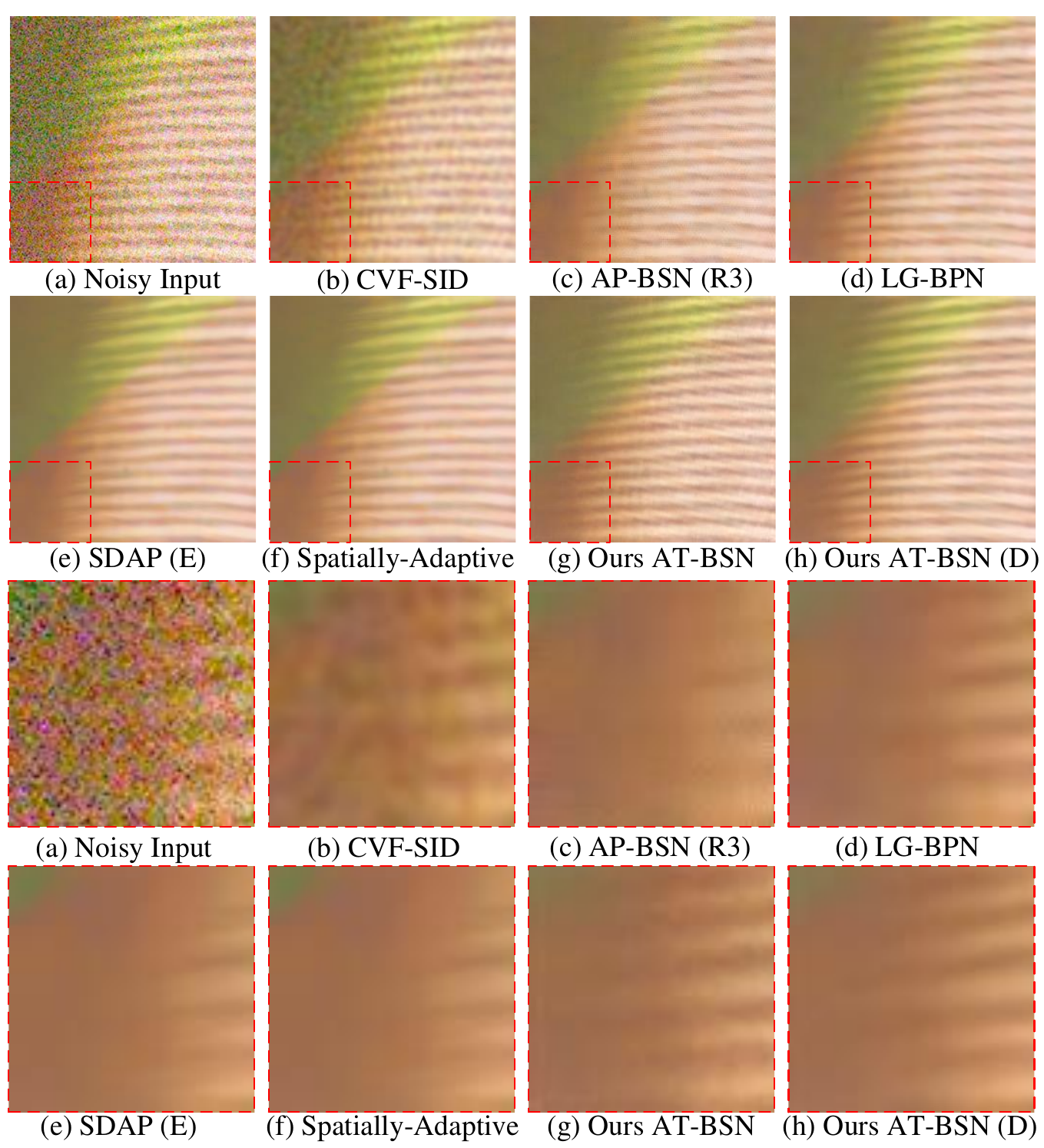}
    \caption{Additional qualitative comparisons on SIDD benchmark dataset.}
    \label{fig:siddbench1}
\end{figure*}

\begin{figure*}[t]
    \centering
    \includegraphics[width=1\linewidth]{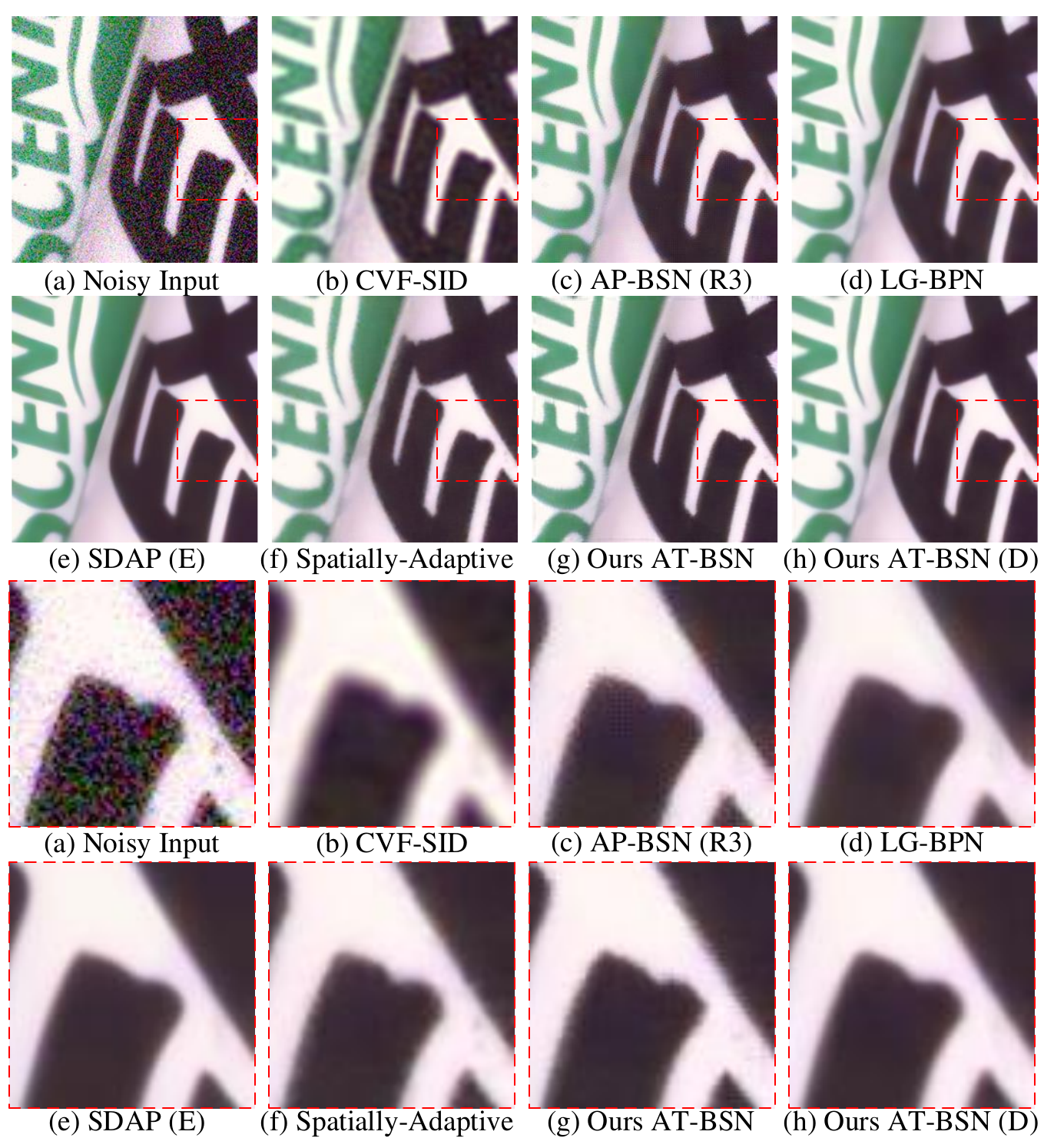}
    \caption{Additional qualitative comparisons on SIDD benchmark dataset.}
    \label{fig:siddbench2}
\end{figure*}

\begin{figure*}[t]
    \centering
    \includegraphics[width=1\linewidth]{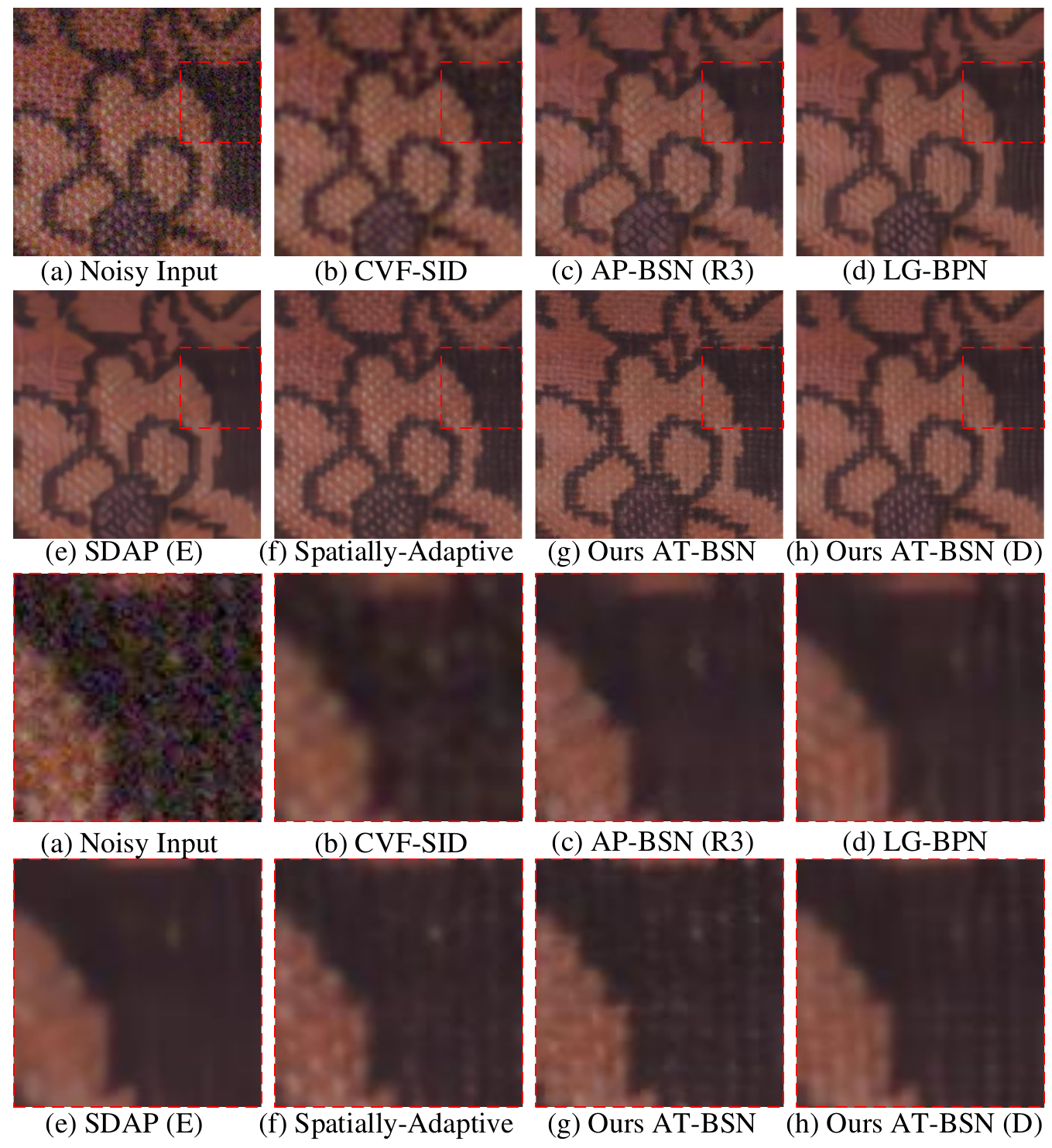}
    \caption{Additional qualitative comparisons on SIDD benchmark dataset.}
    \label{fig:siddbench3}
\end{figure*}

\begin{figure*}[t]
    \centering
    \includegraphics[width=1\linewidth]{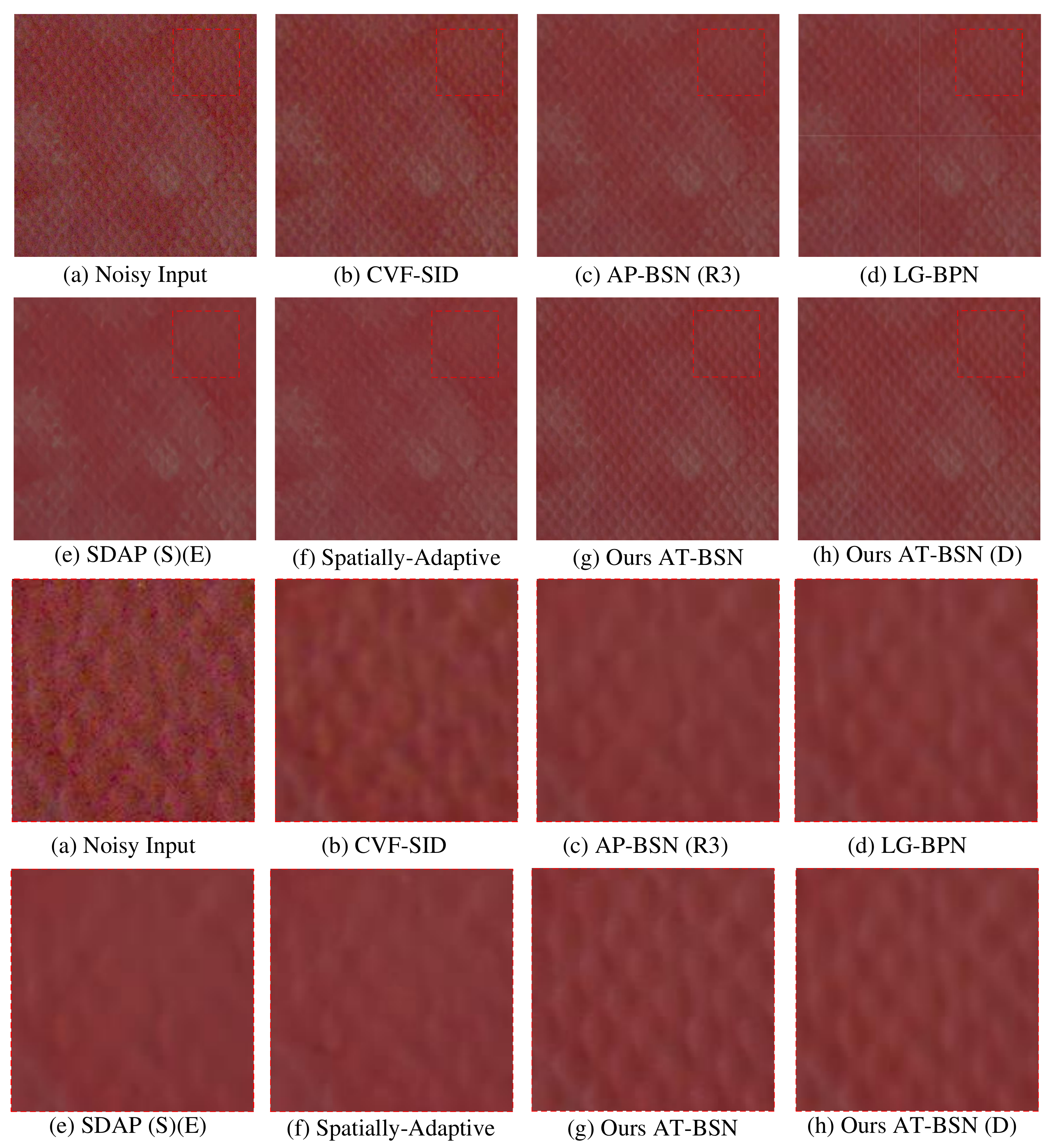}
    \caption{Additional qualitative comparisons on DND benchmark dataset.}
    \label{fig:dndbench1}
\end{figure*}

\begin{figure*}[t]
    \centering
    \includegraphics[width=1\linewidth]{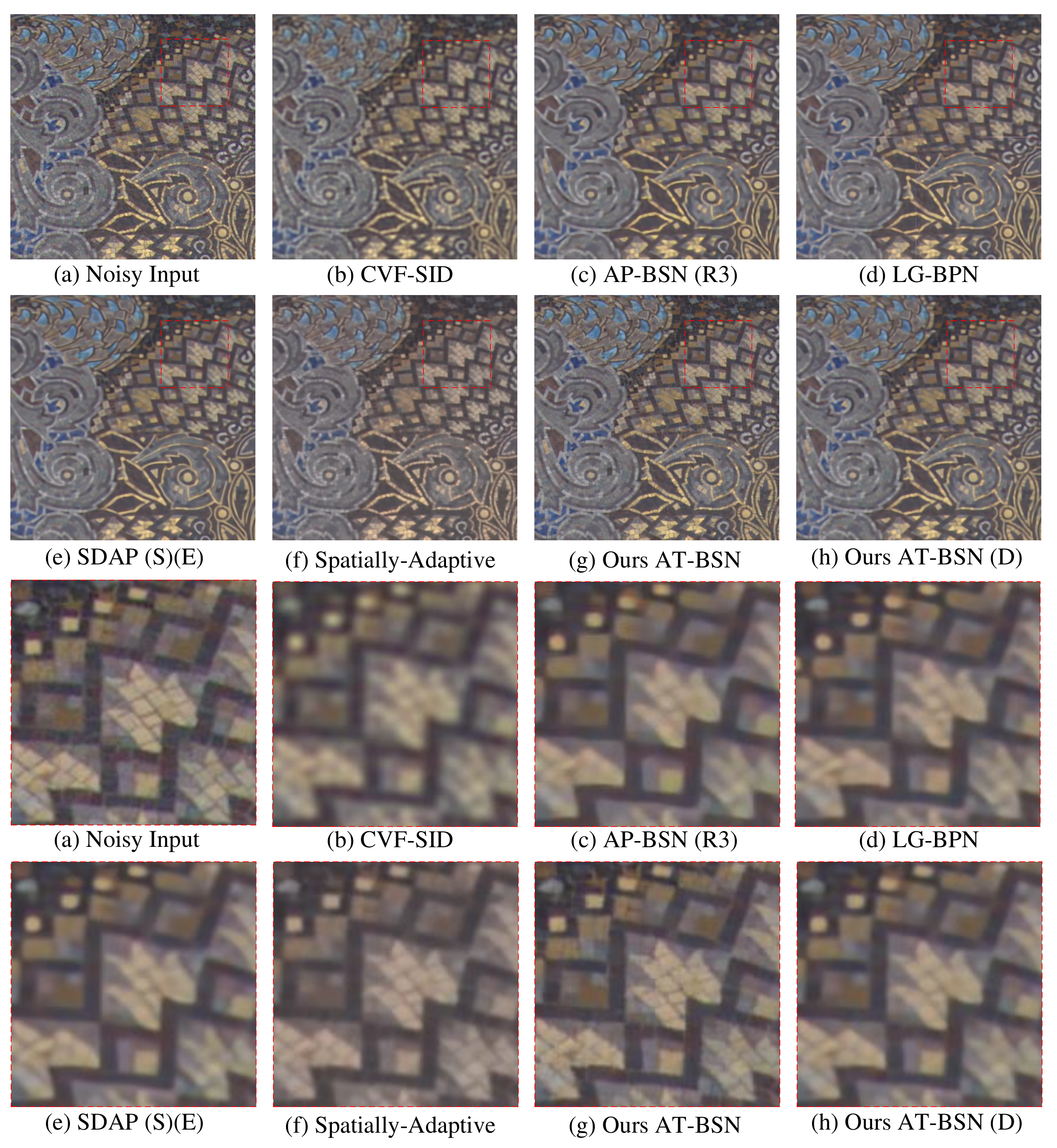}
    \caption{Additional qualitative comparisons on DND benchmark dataset.}
    \label{fig:dndbench2}
\end{figure*}

\begin{figure*}[t]
    \centering
    \includegraphics[width=1\linewidth]{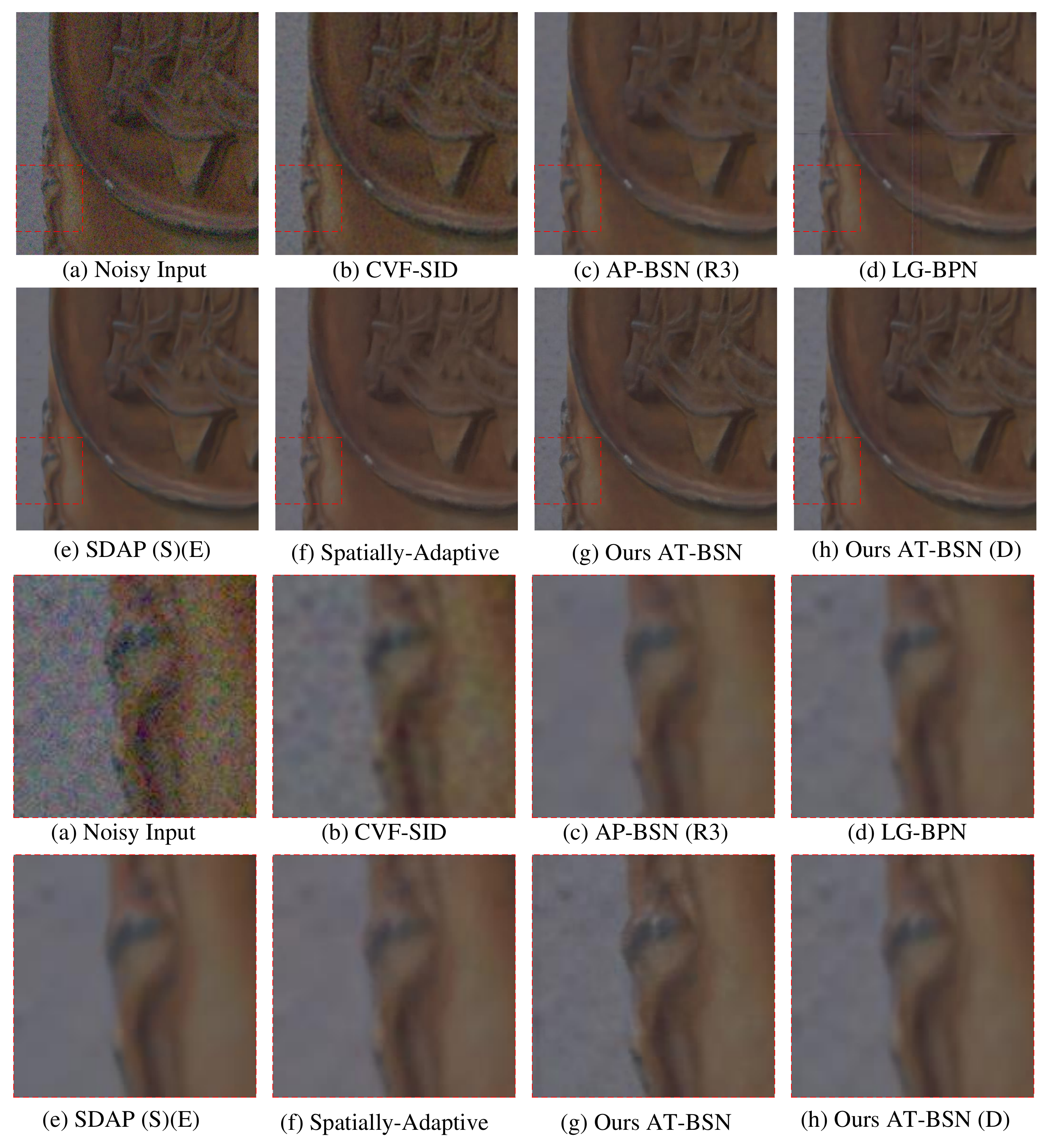}
    \caption{Additional qualitative comparisons on DND benchmark dataset.}
    \label{fig:dndbench3}
\end{figure*}

Fig.~\ref{fig:siddbench1}, Fig.~\ref{fig:siddbench2}, Fig.~\ref{fig:siddbench3} and Fig.~\ref{fig:dndbench1}, Fig.~\ref{fig:dndbench2}, Fig.~\ref{fig:dndbench3} present qualitative comparisons between our proposed method and other approaches on the SIDD and DND benchmark datasets. We apply models trained on the SIDD Medium dataset directly to SIDD benchmark to demonstrate the generalization ability of these methods. For DND dataset, we utilize models trained directly on it to show the advantage of the full self-supervised methods. 

We observe that our method outperforms other methods on both benchmark datasets, producing evidently better denoising results while preserving more texture details and less blur. 
One can find that although AP-BSN(R$^3$)~\cite{APBSN} attempts to use R$^3$~\cite{APBSN} to eliminate the aliasing effect, the aliasing effect cannot be completely eliminated. In addition, although SDAP(E)~\cite{pan2023random} performs well in flat areas, it tends to over-smooth in texture areas, losing many details. SpatiallyAdaptive~\cite{li2023spatially} and LG-BPN~\cite{wang2023lg} can retain more details because they do not downsample the input image. However, the restoration of texture details is still not as good as our method.
We also notice that LG-BPN produces cross artifacts when dealing with larger images in DND datasets ($512\times512$), which we believe is caused by its downsampling operation in the feature domain. It should also be noted that the large kernel operation and post-processing operation of LG-BPN make its computational complexity extremely high, as shown in main paper Tab.4. Our AT-BSN can produce sharper images, while AT-BSN (D) is slightly smoother, achieving a balance between noise removal and texture detail preservation, thus the overall visual effect is better.

